%% file: cocopf.tex
\documentclass{poster14}
\usepackage[utf8]{inputenc}%
\usepackage{amssymb,amsmath}

\usepackage{url}
\usepackage{listings}

\usepackage{graphicx}
\usepackage{xcolor}
\usepackage{rotating}
\usepackage{xstring} 
\usepackage{wasysym} 
\usepackage{MnSymbol} 
\newcommand{\algaperfprof}{Nelder}
\newcommand{\algeperfprof}{L-BFGSB}
\input{bbob_pproc_commands.tex}
\graphicspath{{./}}

\newcommand{\ERT}{\ensuremath{\mathrm{ERT}}}

\newcommand{\Df}{\ensuremath{\Delta f}}

\newcommand{\fopt}{\ensuremath{f_\mathrm{opt}}}
\newcommand{\ftarget}{\ensuremath{f_\mathrm{t}}}

\begin{document}

%
\title{COCOpf: An Algorithm Portfolio Framework}
%
\headtitle{P. BAUDIŠ, COCOpf Algorithm Portfolios}

%
\author{Petr Baudiš}
%
\affiliation{%
	Dept. of Cybernetics, Czech Technical University, Technická 2, 166 27 Praha, Czech Republic
}
  \email{pasky@ucw.cz}

\maketitle


\begin{abstract}
	Algorithm portfolios represent a strategy of composing multiple
	heuristic algorithms, each suited to a different class of problems,
	within a single general solver that will choose the best suited
	algorithm for each input. This approach recently gained popularity
	especially for solving combinatoric problems, but optimization
	applications are still emerging.
	The COCO platform \cite{COCO1} \cite{hansen2012exp} of the BBOB workshop series
	is the current standard way to measure performance of continuous
	black-box optimization algorithms.

	As an extension to the COCO platform, we present
	the Python-based COCOpf framework that allows composing portfolios
	of optimization algorithms and running experiments
	with different selection strategies.
	In our framework, we focus on black-box algorithm portfolio
	and online adaptive selection.  As a demonstration, we measure
	the performance of stock SciPy \cite{SciPy} optimization algorithms and
	the popular CMA algorithm \cite{CMA} alone and in a portfolio with two
	simple selection strategies.  We confirm that even a naive
	selection strategy can provide improved performance across problem classes.
\end{abstract}

\begin{keywords}
	Algorithm portfolio, continuous black-box optimization, hyperheuristics, software engineering.
\end{keywords}


\section{Introduction}

The problem of Algorithm Selection is not new \cite{Rice},
but has only recently gained traction in particular
in the field of combinatoric problem solvers. \cite{combpfsurvey}
There are multiple ways to approach the issue, in our work
we adopt the abstraction of algorithm portfolios. \cite{algportfolios}
The basic idea is that we have a set of
{\em heuristic algorithms} at our disposal (each suitable for
a different class of problem instances), and for each problem
instance on input, we apply a {\em selection strategy} to pick
an algorithm from this portfolio and apply it on the problem.
We can perform the selection once or along a fixed schedule
{\em (offline selection)} based on features of the problem instance,
or in multiple rounds first exploring the suitability of portfolio members,
then allocating time to them based on their previous performance.
In successive rounds, algorithms can be either resumed from their
previous state or restarted; we take the approach of resuming them,
reserving the restart schedule to be an internal matter of each algorithm.%
\footnote{That is, sometimes resumption will simply continue the
	algorithm to a~next iteration, on reaching a local optimum
	the algorithm may attempt to escape it
	and failing that, a complete restart may occur.}

Applying algorithm portfolios on continuous black-box optimization
is still a fresh area of research.%
\footnote{Especially if we do not consider applying related methods
	to individual operator selection within genetic algorithms
	used for optimization.}
The main results so far lie either in population methods, combining
a variety of genetic algorithms together, e.g. the AMALGAM-SO algorithm \cite{amalgam-so},
or in offline methods based chiefly on exploratory landscape analysis \cite{coselsurvey}.

In our work, we aim to allow the more traditional optimization methods
to enter the mix as we regard the algorithms as {\em black-box}, i.e.
we completely avoid modifying their inner working and we simply call
them to retrieve a single result. This allows state-of-art
methods to be combined in a single portfolio easily and offers
easy implementation of whatever selection strategy desired as
the existing algorithm modules can be simply reused without modification.
Furthermore, we focus on online adaptive selection that selects algorithms
based on their performance so far and does not require possibly expensive
feature extraction and training.

The currently accepted de-facto standard for benchmarking optimization
methods is the {\em COmparing Continuous Optimisers} COCO platform \cite{COCO1} \cite{hansen2012exp}
that was originally developed for the BBOB workshop series.
It provides the infrastructure, glue code
for both running experiments and preparing high quality figures,
a set of common reference results and the code for a set of benchmark
functions.  Therefore, we chose to add algorithm portfolio support
to this platform.  The glue code is available in multiple languages,
we extended the Python implementation. We are in part motivated
by the convenient availability of optimization algorithms distributed
along the popular SciPy library \cite{SciPy}.

We have two options when testing algorithm selection methods in practice:
either using only previously published data on preformance of individual algorithms,
or re-running optimization algorithms from scratch under the guide of
a particular selection method.  The former approach is certainly an easier
option while benchmarking, nevertheless we opted for the latter model so that
we can also readily apply all code on practical function optimization
aside of mere benchmarking.  Apart of that, the available performance data
does not include information about algorithm iterations, listingonly individual
function evaluations. That makes it unclear when is a good time to switch.
Lastly, the re-running model allows migration of intermediate solutions
between algorithms as a simple extension.

In Section \ref{framework}, we introduce the COCOpf software, explaining its
architecture, general API and a few interesting implementation points.
In Section \ref{benchmark}, we establish a reference algorithm portfolio,
benchmarking its individual members and two algorithm selection methods
implemented in the framework. We present the benchmark results in Section
\ref{results} and conclude with a results summary and outline of
a few directions of future work in Section \ref{conclusion}.

\section{COCOpf Framework}
\label{framework}

Our framework has been implemented as a Python module that can be
imported by user applications. The framework provides the means to run
the experiment on a given set of benchmark scenarios, maintains
a population of algorithm instances and provides a complex wrapper to
individual algorithms that uses an existing per-iteration callback
mechanism to suspend and resume execution of the algorithm after
a single iteration.
Algorithm wrappers for the SciPy optimization methods and the
reference implementation of CMA are available out-of-the box
as well as two example algorithm selection strategies that are
benchmarked below.
The framework is publicly available as free software under the
permissible MIT licence at \url{https://github.com/pasky/cocopf}.

In design of such a framework, a choice presents itself --- to either
build a generic main program which allows an experiment to choose
a certain configuration and plug in custom callbacks, or to prepare
a set of "lego bricks" that replace common parts of the main program
but letting each experiment provide a separate main program.  We opted
for the latter; it reduces encapsulation and requires slightly more
repetitive code, but it allows a much greater degree of flexibility
encouraging diverse experiments.

\subsection{Experiment Wrapper}

The {\bf Experiment} class wraps up the common code that sets up
the COCO benchmark, initializing the logs, sets of dimensions
and function instances etc. It also provides an iterator that
in turn returns individual function instances to be minimized,
and a pretty-printing progress report function. After an experiment
is finished, the standard BBOB postprocessing tools may be used
to examine and plot the performance data. An example of usage
is shown in Fig. \ref{figmain}.


\begin{figure}[ht!]
\begin{center}
\begin{lstlisting}
e = Experiment(maxfev=10000, shortname="EG50",
	       comments="Eps-greedy, eps=0.5")
for i in e.finstances():
    minimization(i)
    e.f.finalizerun()
    e.freport(i)
\end{lstlisting}
\caption{A minimalistic main body of a Python script using the Experiment class. Function {\em minimization} is to be provided by the user.}
\label{figmain}
\end{center}
\end{figure}

\subsection{Minimization Wrappers}

All algorithms are executed via the {\bf MinimizeMethod} class.
Its primary purpose is to produce a callable Python object from
a user-provided string name of algorithm.
By default, it calls the {\em scipy.optimize.minimize} API with
unchanged method name and wraps it in the {\em scipy.optimize.basinhopping} API.
However, it is expected the user will create a subclass adding custom optimizers;
a~demo subclass adding the CMA optimization is included.%
\footnote{The custom optimizers may also legitimately want
to use the Basin Hopping algorithm (see Sec. \ref{bhop}),
but SciPy allows using custom minimization routines with the 
basinhopping API only starting with version 0.14.}

We were presented with a software engineering problem of how
to suspend, resume and cancel the algorithms between iterations
without requiring any modification to the algorithm code,
while all we have is a callback executed after each iteration.
Our answer is the {\bf MinimizeStepping} class which creates
a separate thread for each algorithm instance and uses the
standard Python {\em Queue} mechanism to pause execution of
each thread within the callback code based on instructions
from the main thread. At any time, only a single thread is running.
To cancel execution (when time is up or a solution is found)
we pass a special message through the queue that raises an
exception within the callback, eventually cancelling the whole
algorithm and thread orderly.
An ``mlog'' facility is provided to log function evaluation status
at the end of each algorithm iteration.
\label{mlog}

\subsection{Portfolio Population}

The {\bf Population} class maintains a population of algorithms
drawn from the algorithm portfolio; each population member
corresponds to a tuple of method, solution coordinates, current value
and number of iterations invested.  Usually, the size of population
is static, but dynamically adding new members is also supported.

In a typical case, the population
size is the same as the portfolio size, meaning that only a single
solution is tracked for each algorithm.  However, in principle this
may not be the case.  In fact, we can also make a population from
solutions all using a single algorithm --- this special case corresponds
to the problem of finding an efficient restart or resume schedule
for a particular algorithm.

\section{Benchmarking Simple Portfolios}
\label{benchmark}

To showcase COCOpf, we composed a reference portfolio from
readily available open source Python optimizers and implemented two
selection strategies.

\subsection{Reference Algorithm Portfolio}

As a testbed algorithm portfolio, we have chosen the six stock minimizers provided
by SciPy v0.13 \cite{SciPy} that are available for direct use:%
\footnote{Other minimizers are also available in SciPy,
	but they are either unsuitable for black-box minimization or do
	not provide a callback mechanism to allow pausing execution between
	individual iterations.}

\begin{itemize}
	\item {\bf Nelder-Mead} uses the Simplex algorithm. \cite{NM1} \cite{NM2}
	\item {\bf Powell} implements a tweaked version of the Powell's conjugate direction method. \cite{Powell1}
	\item {\bf CG} is a nonlinear conjugate gradient method using the Fletcher-Reeves method. \cite[p.~120--122]{NO}
	\item {\bf BFGS} uses the quasi-Newton method of Broyden, Fletcher, Goldfarb and Shanno \cite[p.136]{NO} with numerically approximated derivations.
	\item {\bf L-BFGS-B} is a limited-memory variant of BFGS with box constraints. \cite{LBFGSB1} \cite{LBFGSB2}
	\item {\bf SLSQP} implements Sequential Least SQuares Programming \cite{SLSQP} with inequalities as the box constraints.
\end{itemize}

All of these SciPy minimizers perform local minimization; to achieve global
optimization, we wrap them in the ``Basin Hopping'' restart strategy
\cite{basinhopping} (also provided by SciPy), which is conceptually similar
to Simulated Annealing with a fixed temperature. The strategy performs
the hopping step 100 times (the default) before restarting from scratch.
On restart, the starting position is determined uniformly randomly
in the $[-5,5]^k$ space.
\label{bhop}

To improve the portfolio performance on more difficult functions,
we also included the popular CMA algorithm \cite{CMA}
(in the ``production`` version of its official reference
Python implementation), i.e. a genetic algorithm
that follows the Covariance Matrix Adaptation evolution strategy.

In accordance with our black-box approach, we kept all the algorithms
at their default settings and did not tune any of their parameters.
Where applicable, we set the box constraints to $[-6,6]^k$ bounds.%
\footnote{The optimum is guaranteed to be in $[-5,5]^k$,
but some minimizers may behave erraticaly when the optimum is near to or at the bounds.}

Each algorithm provides a callback method that is invoked after
a single iteration of the algorithm; we suspend the algorithm
after each callback invocation.

The reference portfolio is not perfectly balanced and certainly
would benefit from algorithms with more diverse performance.
Our motivation for going with these is that the implementations
are free software, trivial to obtain and therefore representing
a very low barrier of entry.  Also, a~SciPy-based portfolio
gives valuable feedback to SciPy users regarding the best algorithm
to use on a certain function, and provides a gateway to introduce
an algorithm portfolio selection strategy to SciPy itself in the future.

\subsection{Algorithm Selection Strategies}

While the core of our followup work involves researching various
algorithm selection strategies, in the context of presentation
of the COCOpf framework, we decided to demonstrate just the
performance of two simple simplest methods whose source code
is also a part of the framework:

\begin{itemize}
	\item The {\bf UNIF} strategy performs a number of rounds
		where in every round, a uniformly randomly selected
		algorithm is run for a single iteration.
	\item The {\bf EG50} strategy performs a number of rounds
		where in every round, an {\em epsilon-greedy policy}
		with $\varepsilon=0.5$ selects the algorithm to run for
		one iteration.  That is, the algorithm with
		the currently best solution is run with $p=0.5$ and
		a randomly chosen algorithm is run otherwise.
\end{itemize}

\section{Results}
\label{results}

Results from experiments according to \cite{hansen2012exp} on the
benchmark functions given in \cite{wp200901_2010,hansen2012fun} are
presented in Figures~2, 3 and 4.
\footnote{The full results dataset is available at \url{http://pasky.or.cz/sci/cocopf-scipy}.}
The \textbf{expected running time (ERT)} used in the figures
depends on a given target function value, $\ftarget=\fopt+\Df$, and is
computed over all relevant trials as the number of function
evaluations executed during each trial while the best function value
did not reach \ftarget, summed over all trials and divided by the
number of trials that actually reached \ftarget\
\cite{hansen2012exp,price1997dev}.


If we review the results in context of choosing a best (black-box function)
SciPy optimizer, we can see that there is no single right choice of a SciPy optimizer
that works on all function instances and computational budgets. We can
observe that {\bf Powell} generally works best for separable functions while
{\bf BFGS} and {\bf SLSQP} are good choices for many
functions, especially in high dimensions. For non-separable functions,
{\bf CMA} sooner or later overtakes the SciPy optimizers but often with
a slow start (which is relevant in case a function evaluation is very expensive)
and is not competitive with the SciPy optimizers on separable
functions.

\begin{figure*}
\centering
\begin{tabular}{@{}c@{}c@{}c@{}c@{}}
\includegraphics[width=0.253\textwidth, trim= 0.7cm 0.8cm 0.5cm 0.5cm, clip]{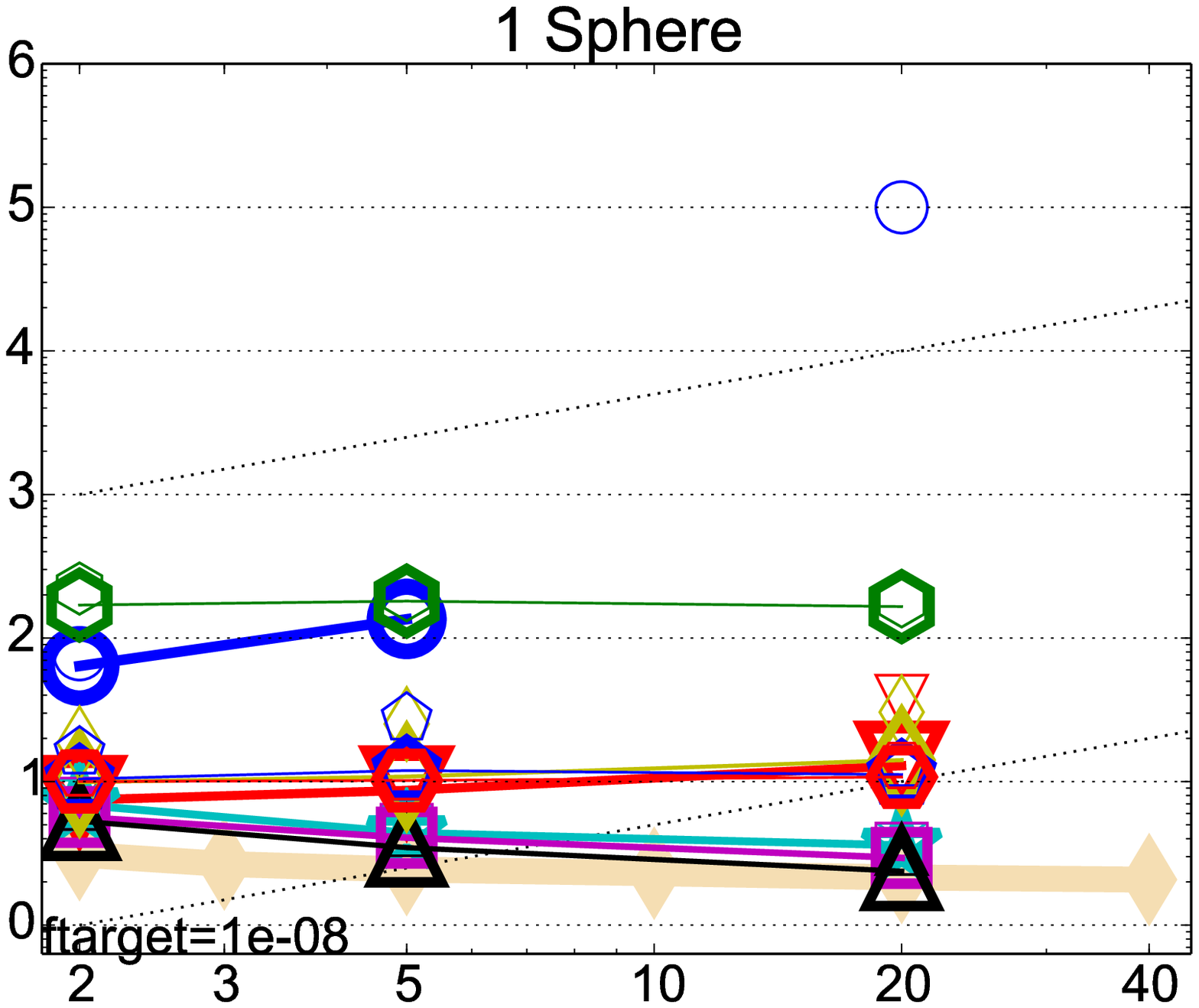}&
\includegraphics[width=0.238\textwidth, trim= 1.8cm 0.8cm 0.5cm 0.5cm, clip]{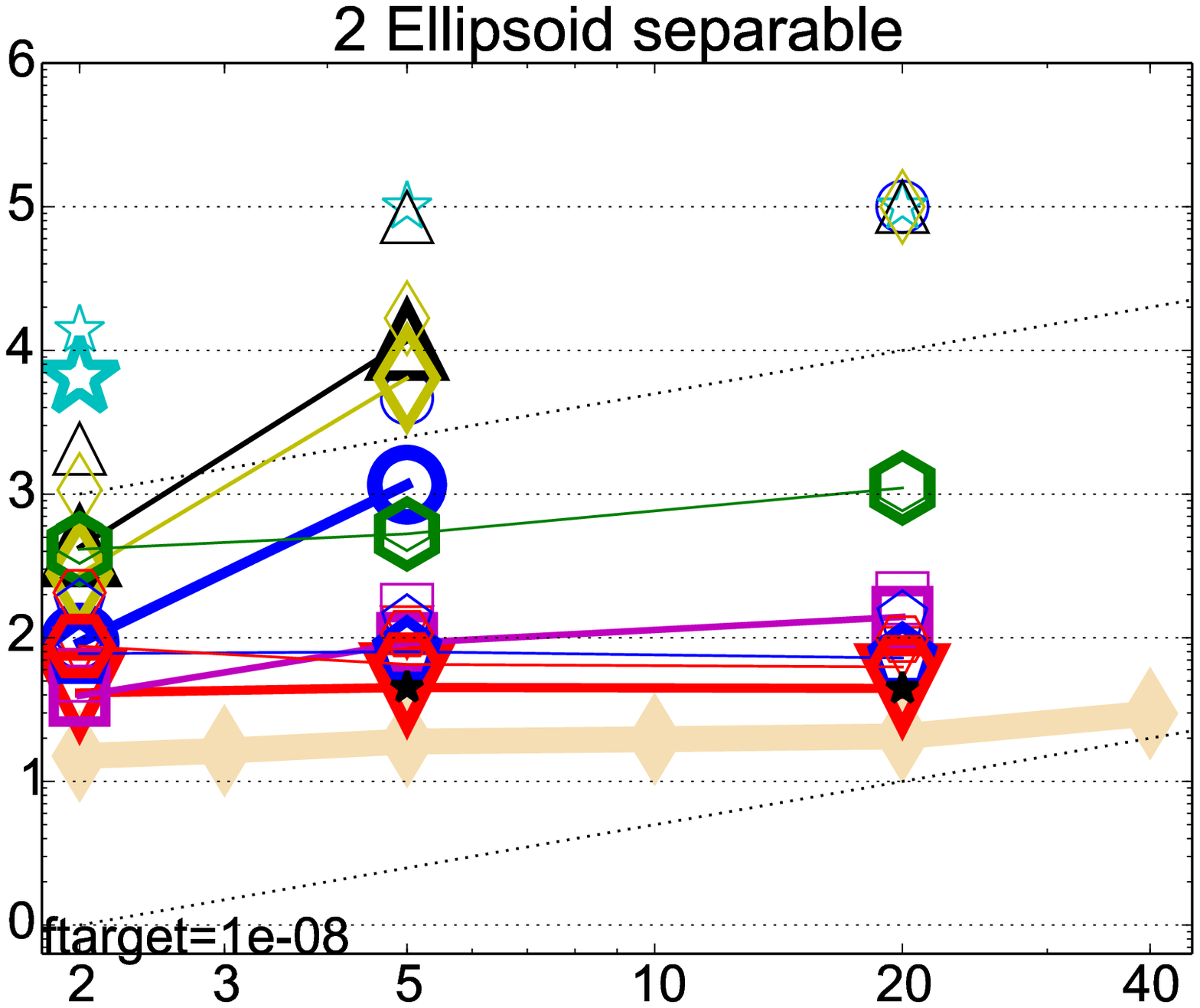}&
\includegraphics[width=0.238\textwidth, trim= 1.8cm 0.8cm 0.5cm 0.5cm, clip]{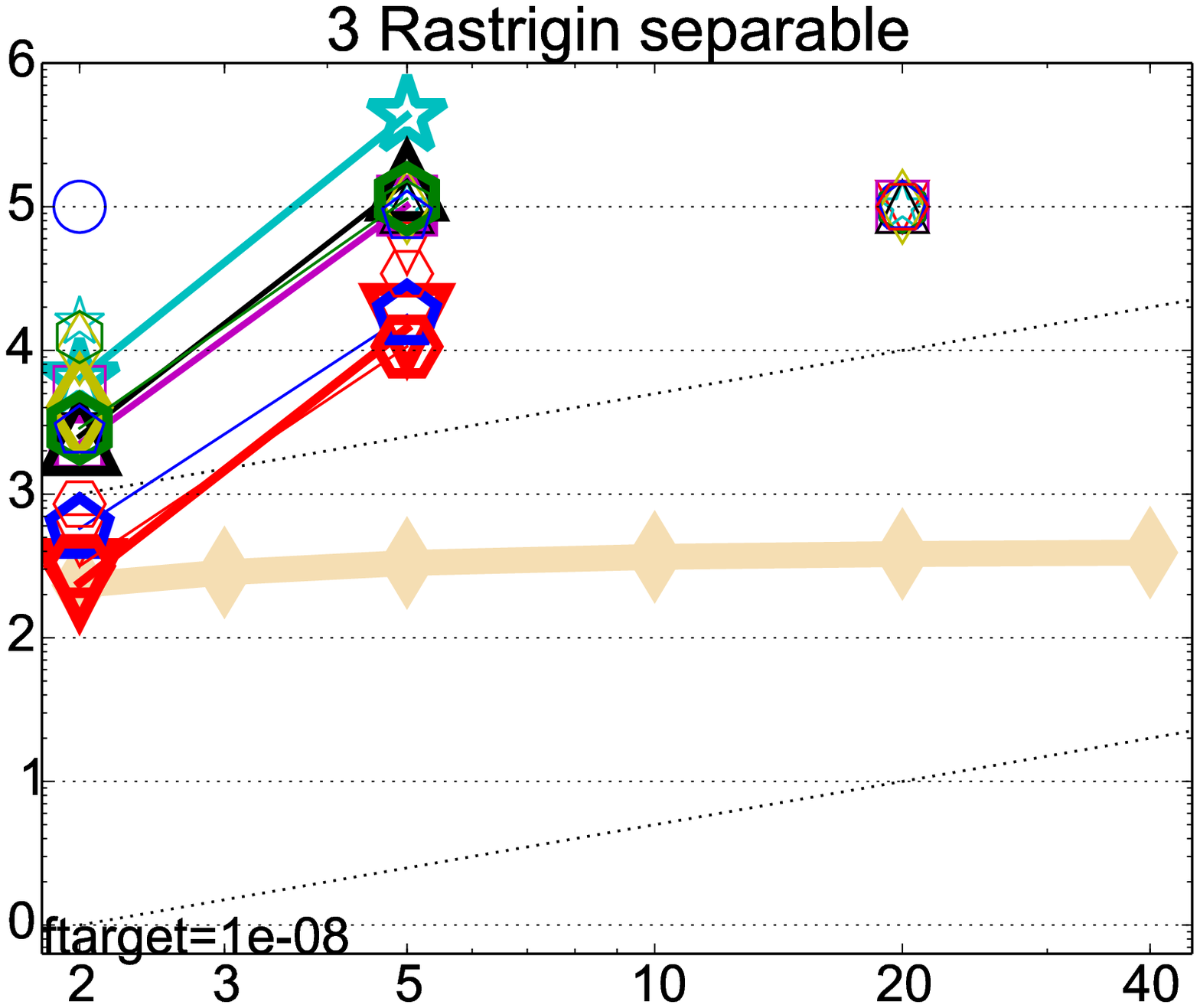}&
\includegraphics[width=0.238\textwidth, trim= 1.8cm 0.8cm 0.5cm 0.5cm, clip]{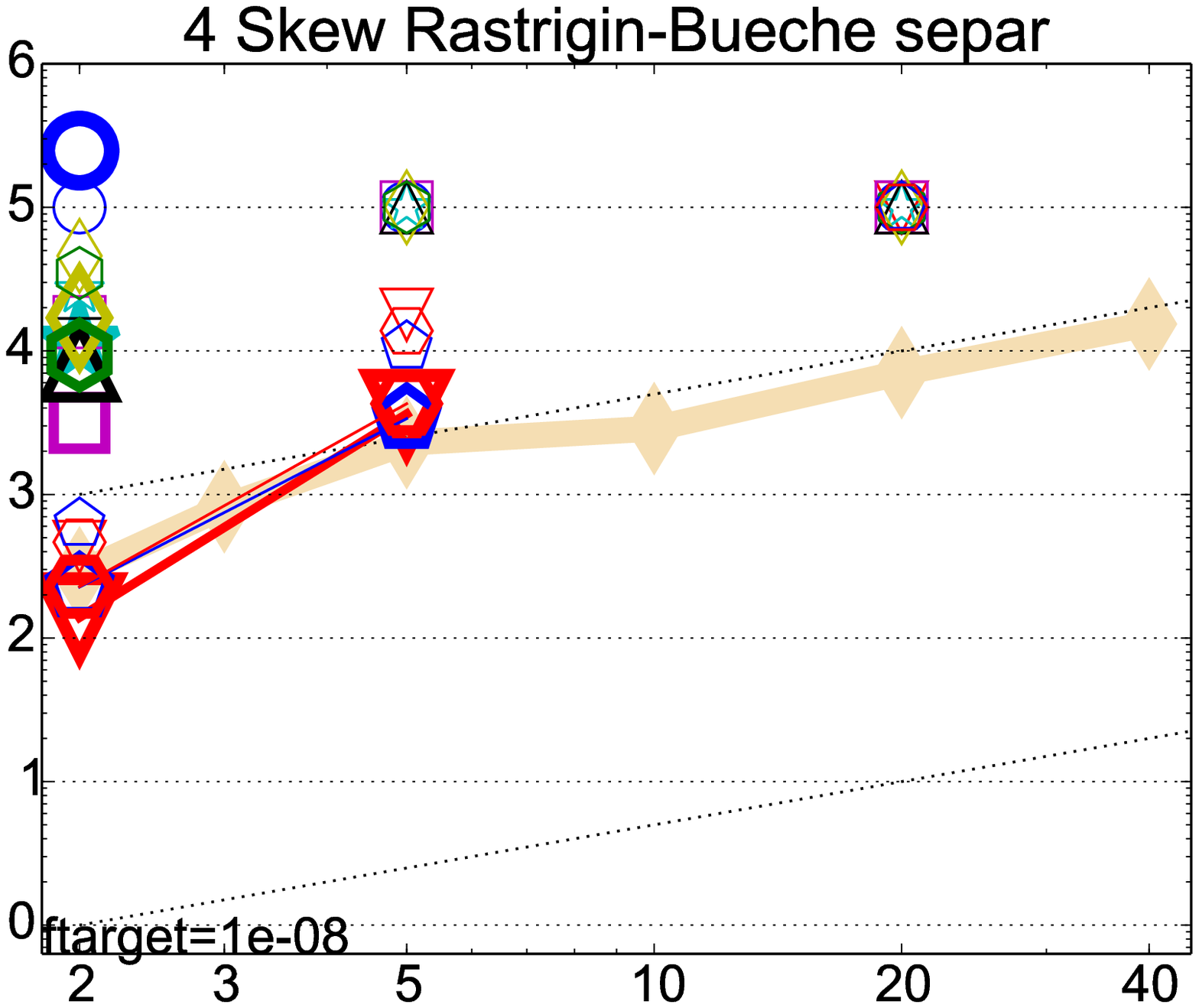}\\
\includegraphics[width=0.253\textwidth, trim= 0.7cm 0.8cm 0.5cm 0.5cm, clip]{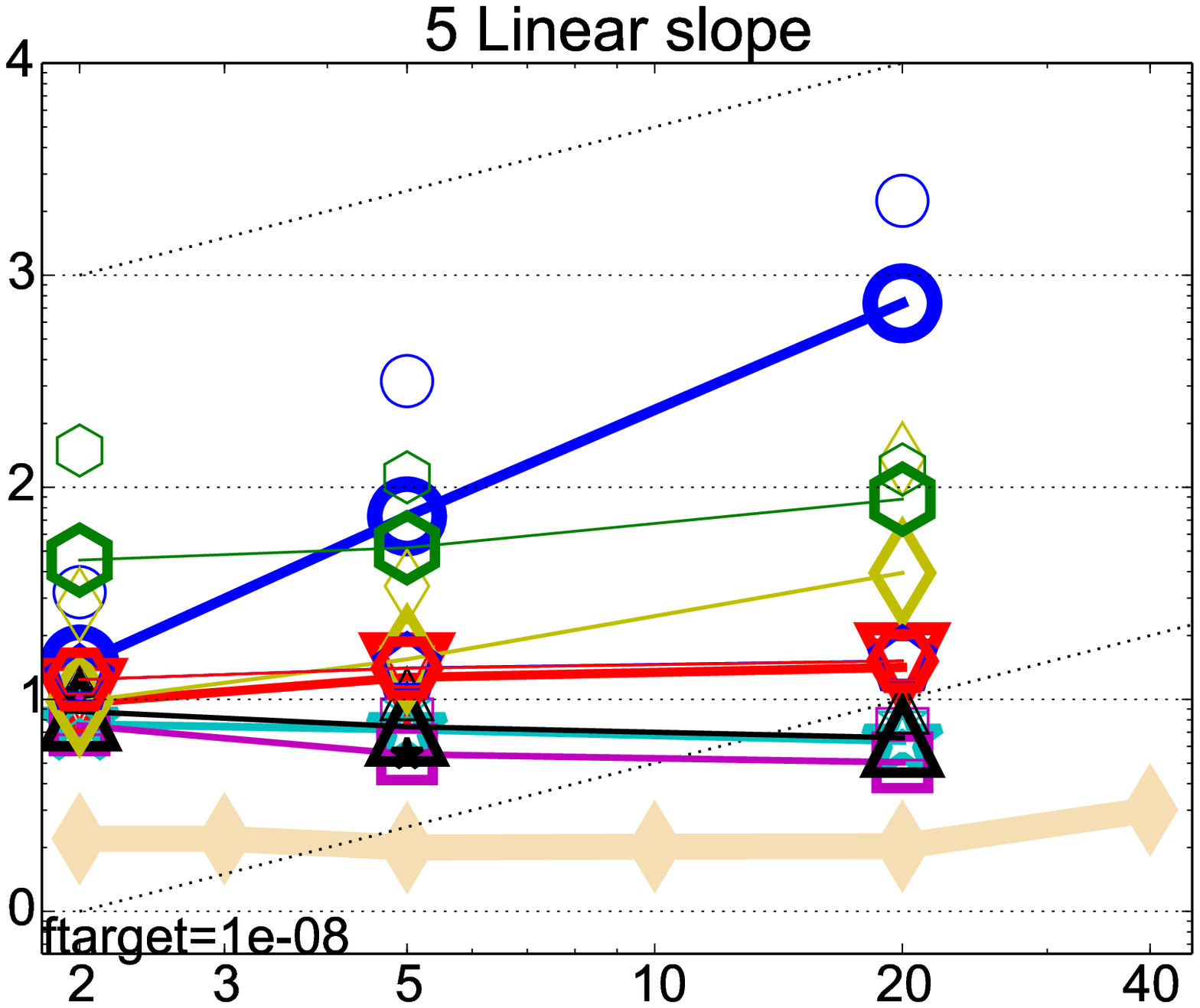}&
\includegraphics[width=0.238\textwidth, trim= 1.8cm 0.8cm 0.5cm 0.5cm, clip]{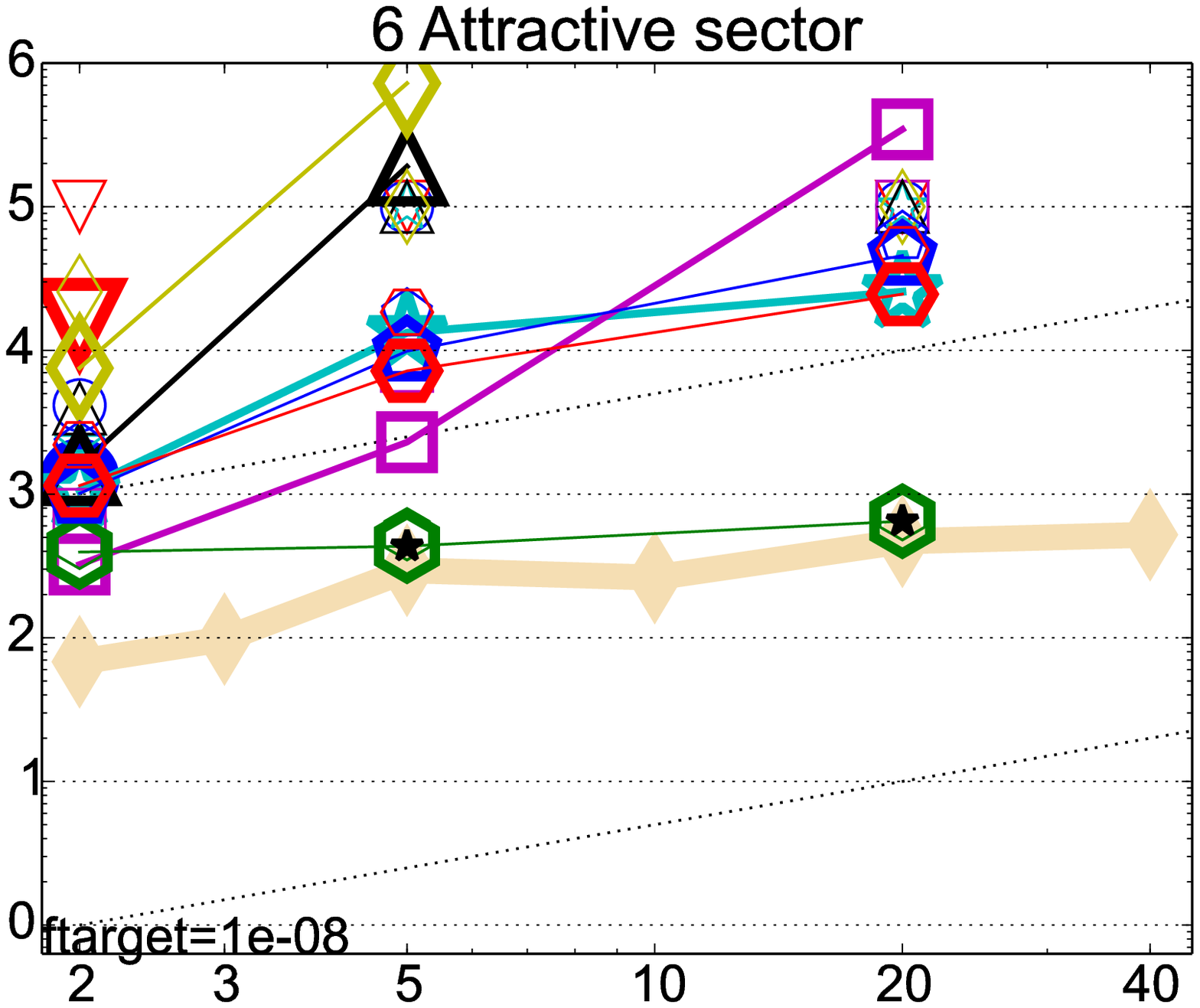}&
\includegraphics[width=0.238\textwidth, trim= 1.8cm 0.8cm 0.5cm 0.5cm, clip]{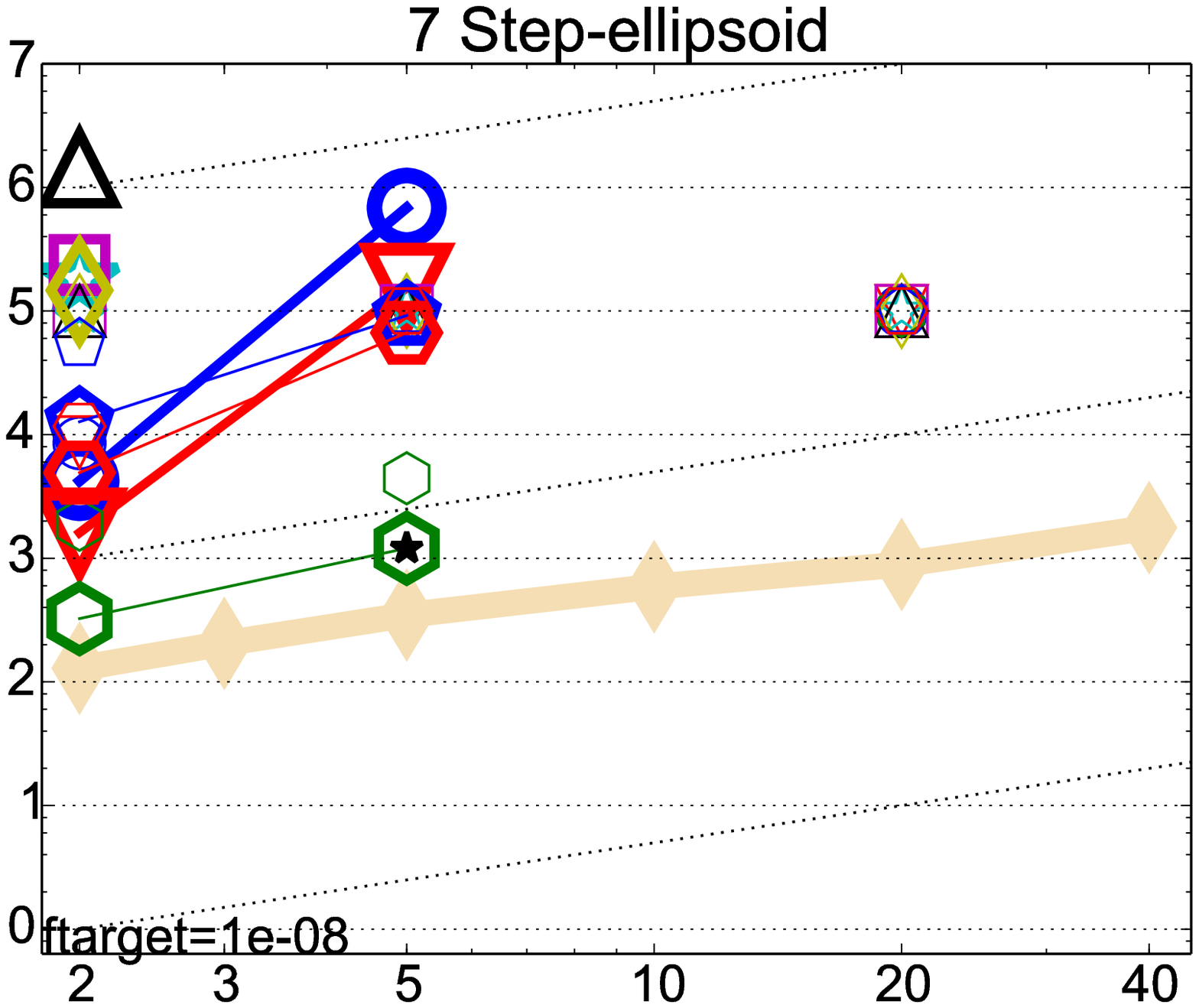}&
\includegraphics[width=0.238\textwidth, trim= 1.8cm 0.8cm 0.5cm 0.5cm, clip]{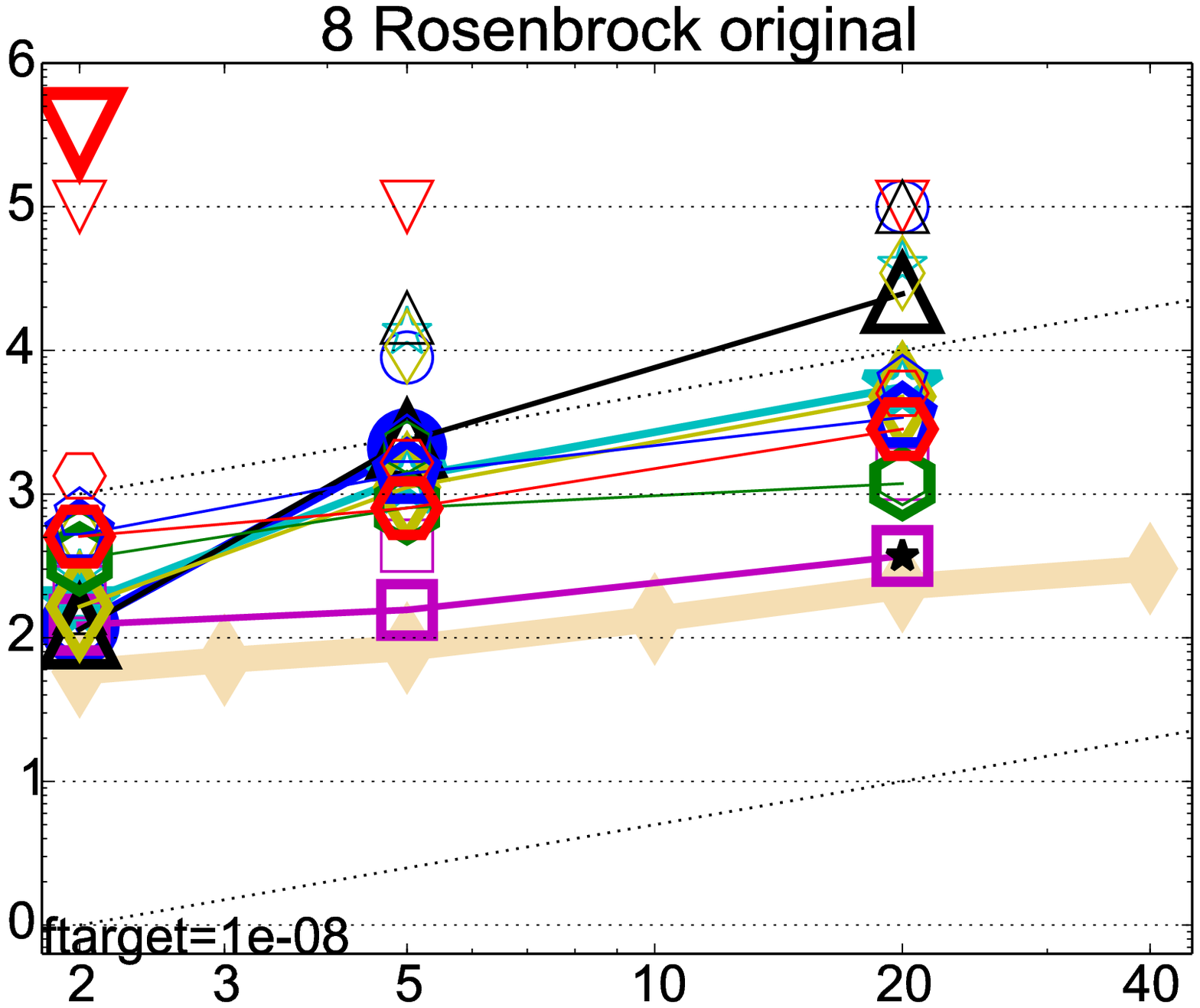}\\
\includegraphics[width=0.253\textwidth, trim= 0.7cm 0.8cm 0.5cm 0.5cm, clip]{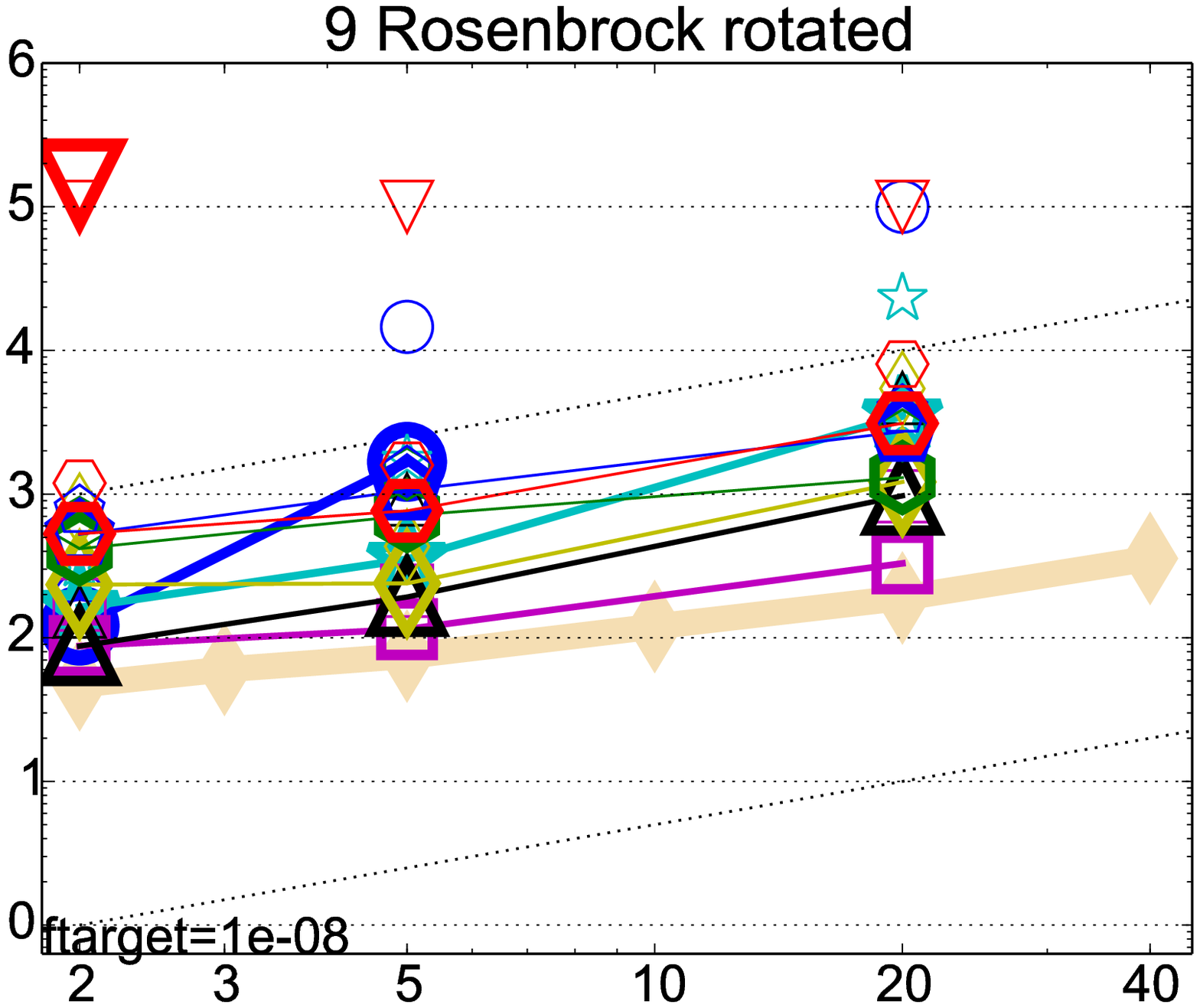}&
\includegraphics[width=0.238\textwidth, trim= 1.8cm 0.8cm 0.5cm 0.5cm, clip]{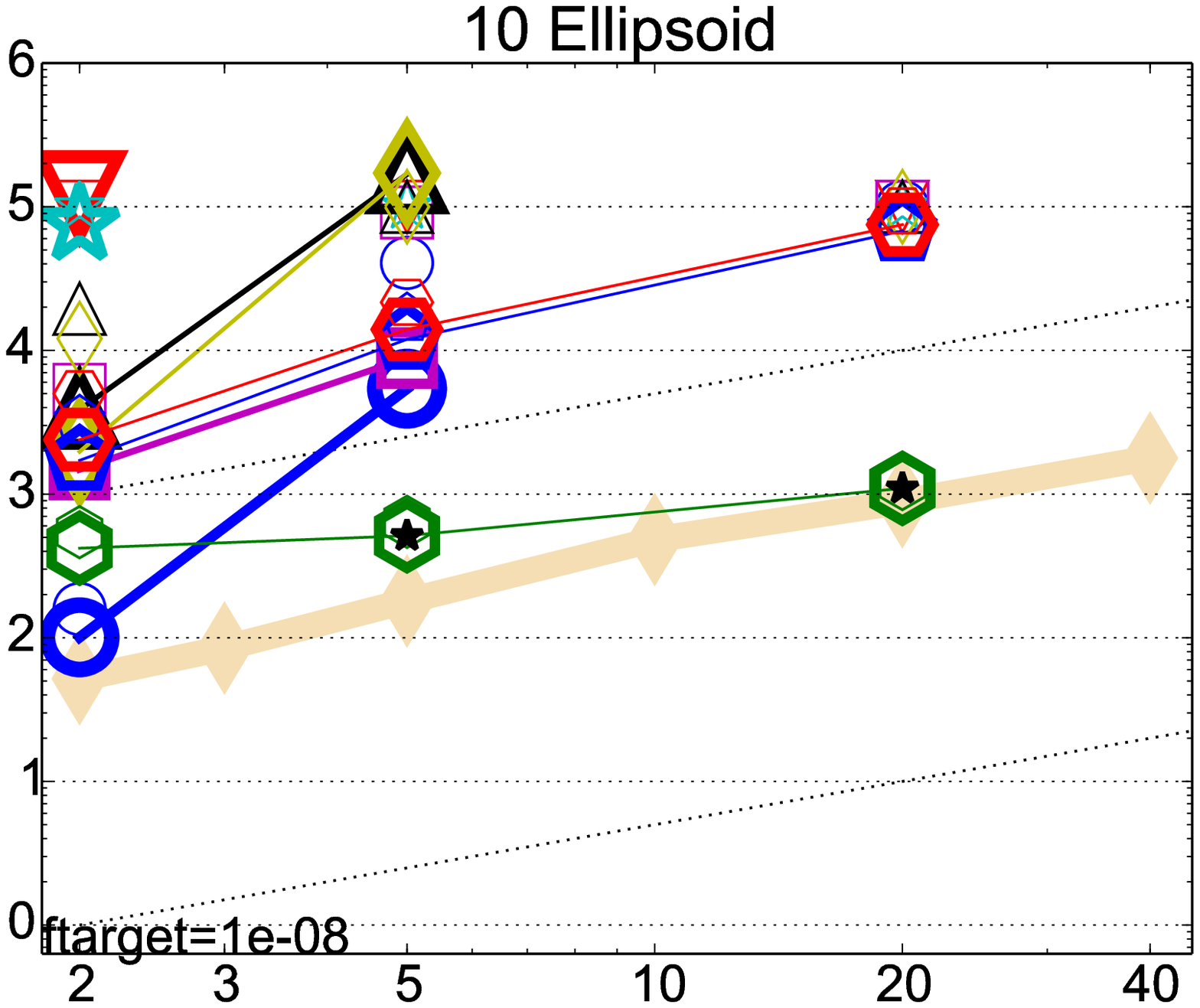}&
\includegraphics[width=0.238\textwidth, trim= 1.8cm 0.8cm 0.5cm 0.5cm, clip]{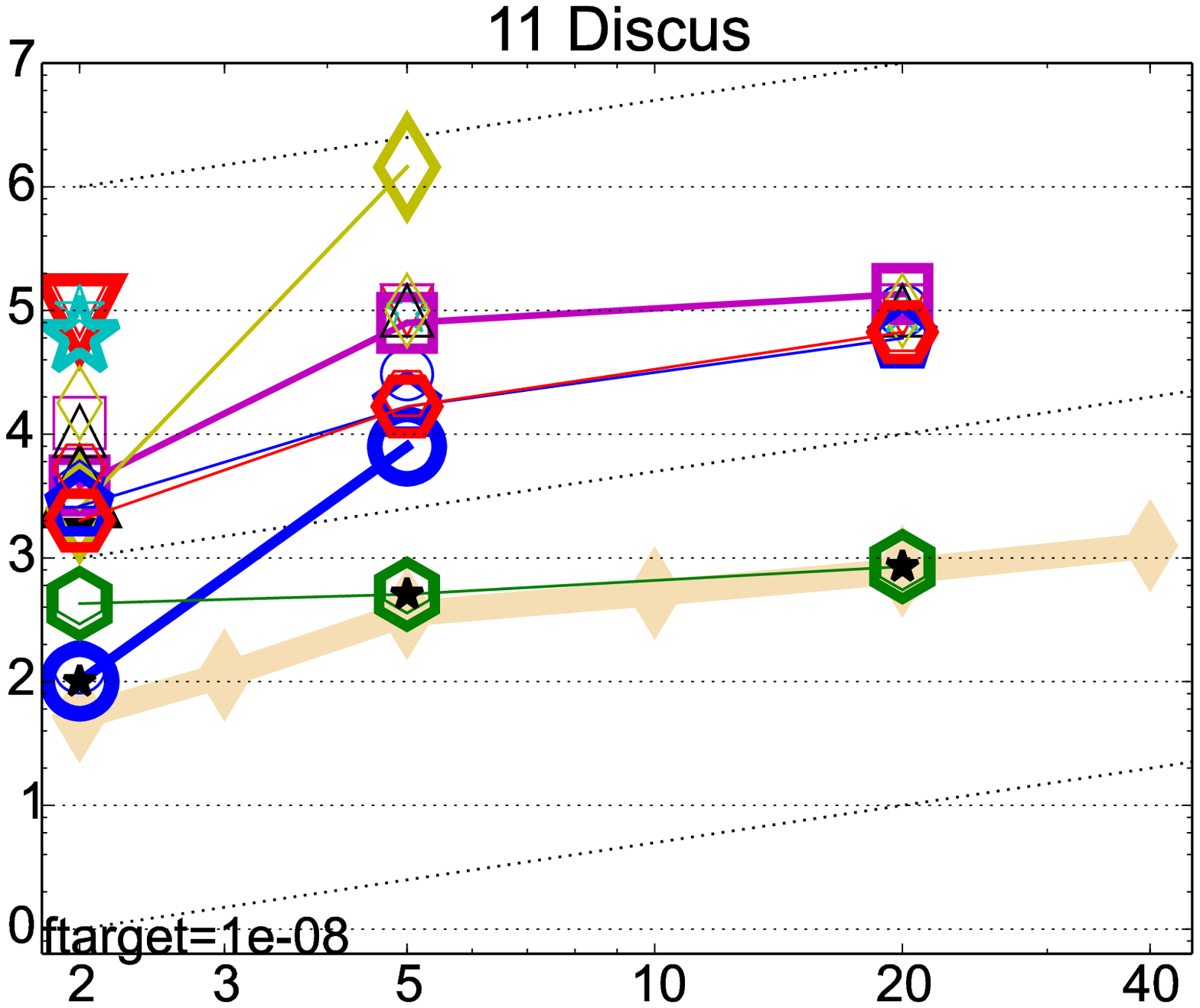}&
\includegraphics[width=0.238\textwidth, trim= 1.8cm 0.8cm 0.5cm 0.5cm, clip]{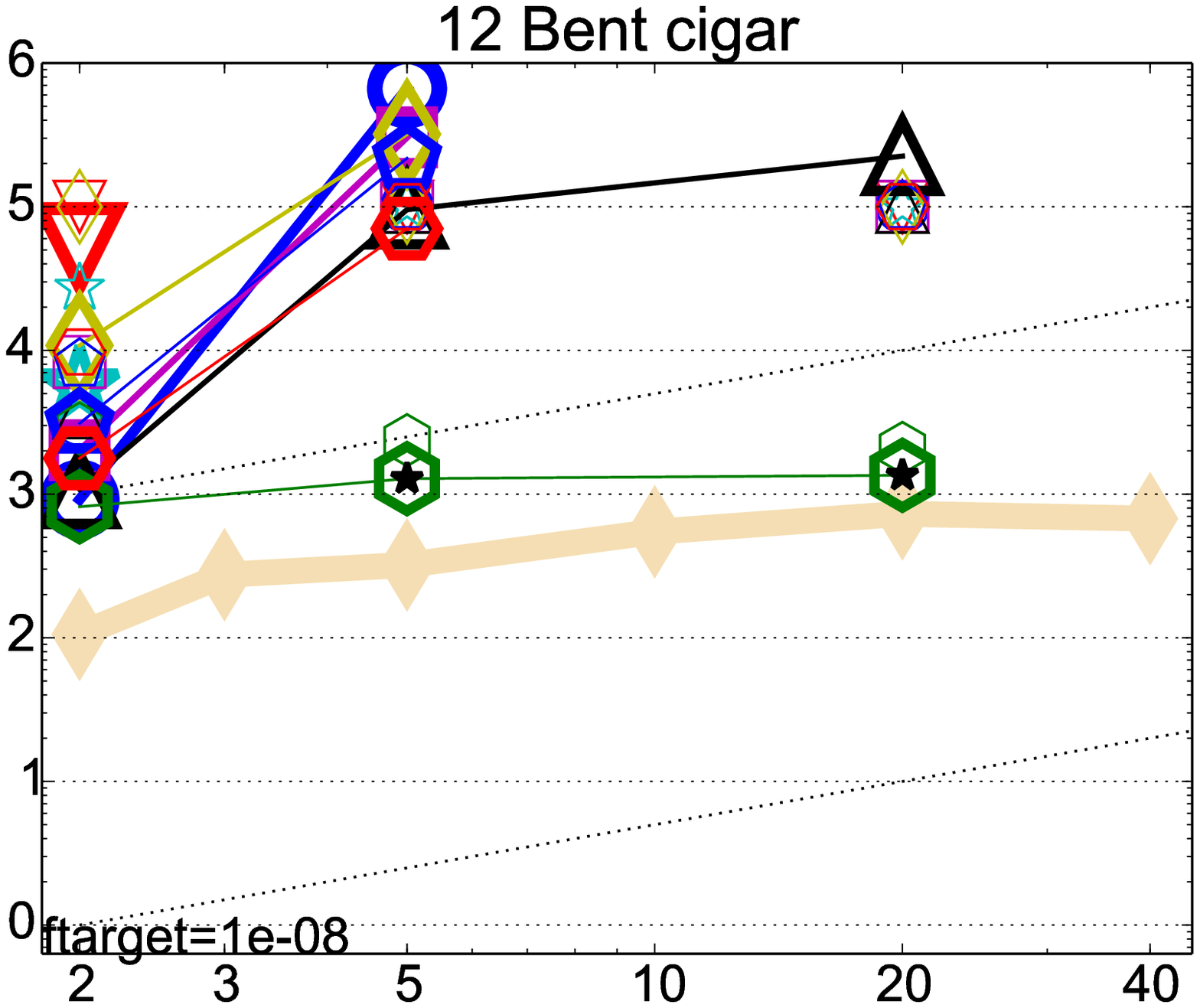}\\
\includegraphics[width=0.253\textwidth, trim= 0.7cm 0.8cm 0.5cm 0.5cm, clip]{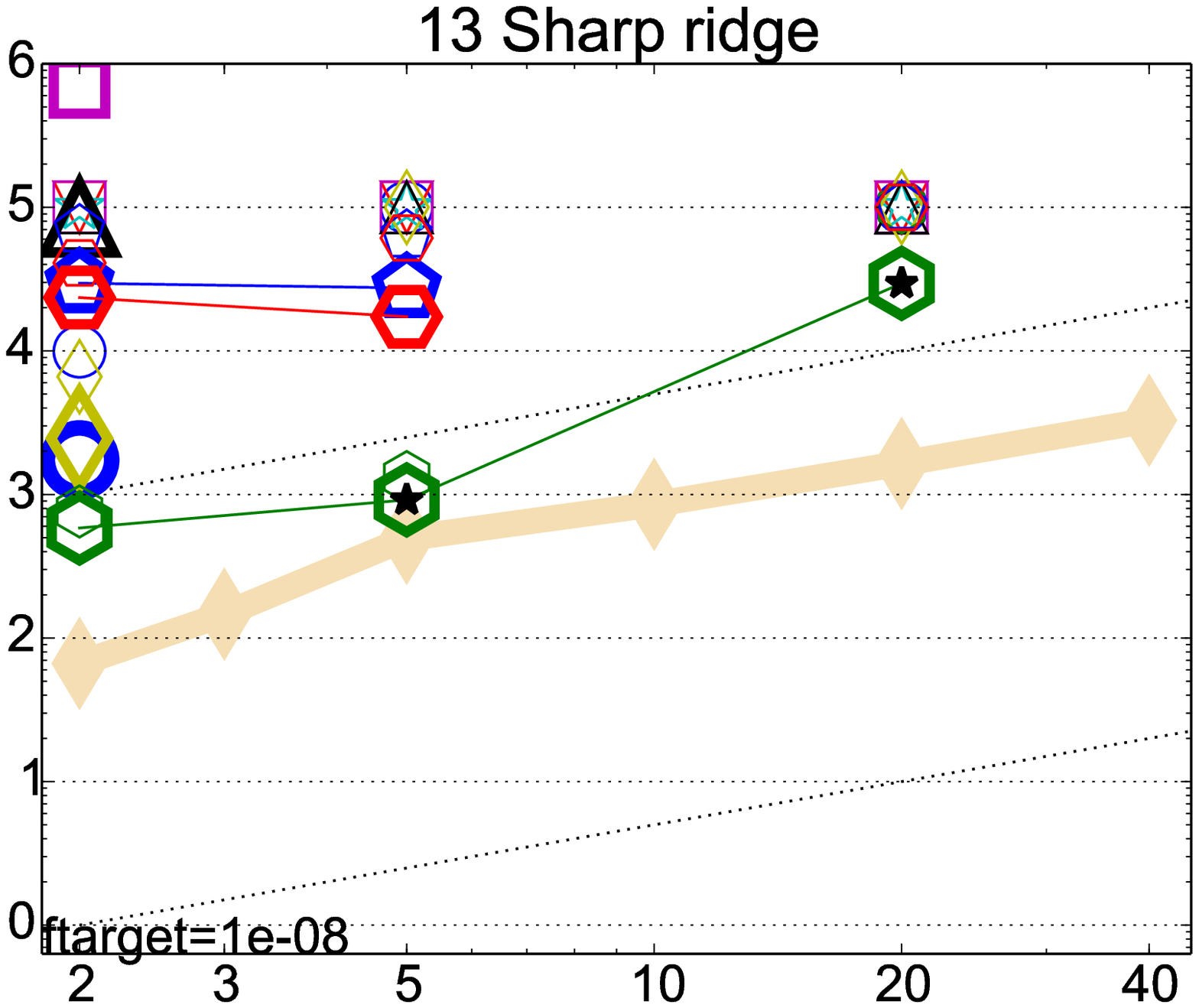}&
\includegraphics[width=0.238\textwidth, trim= 1.8cm 0.8cm 0.5cm 0.5cm, clip]{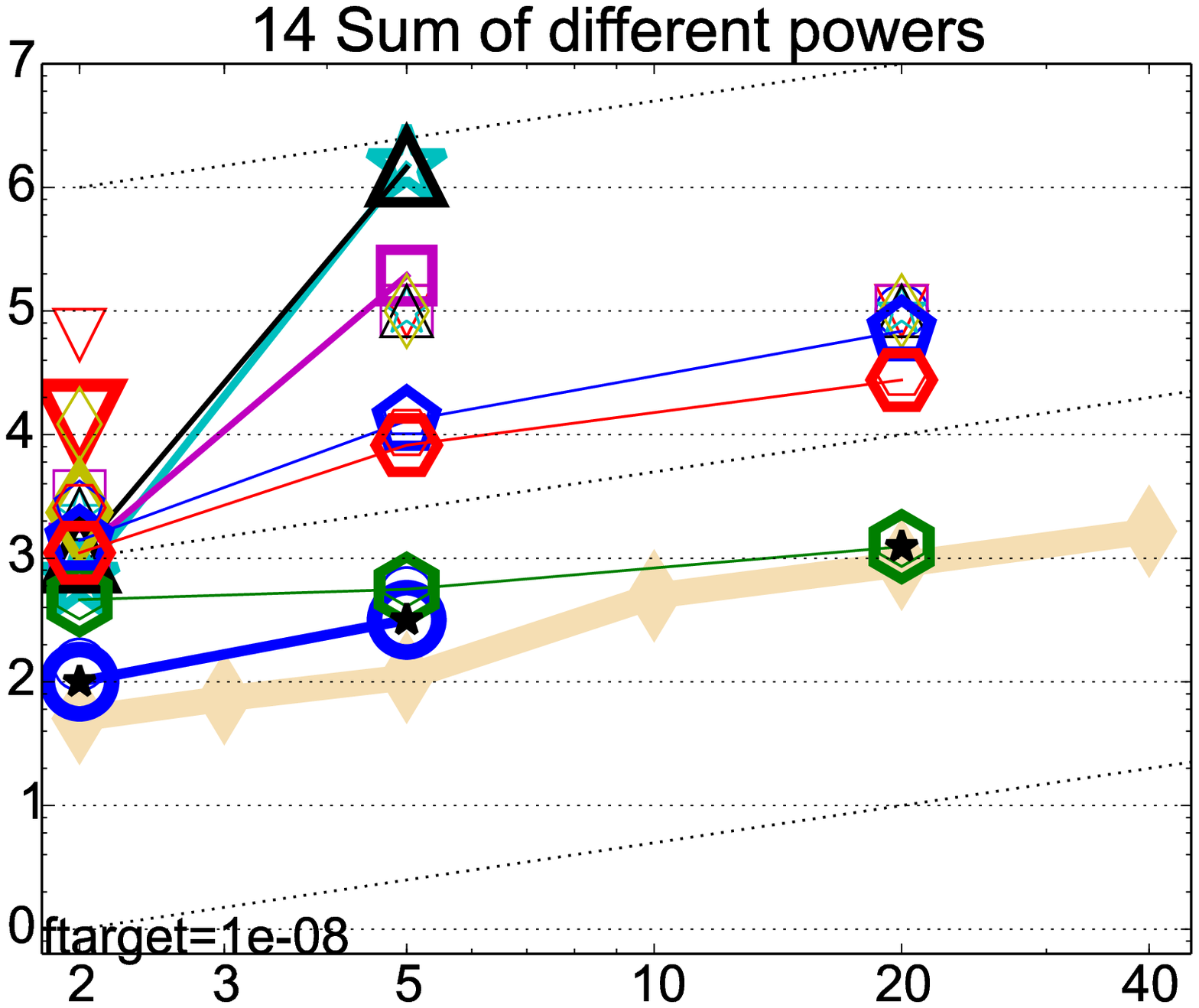}&
\includegraphics[width=0.238\textwidth, trim= 1.8cm 0.8cm 0.5cm 0.5cm, clip]{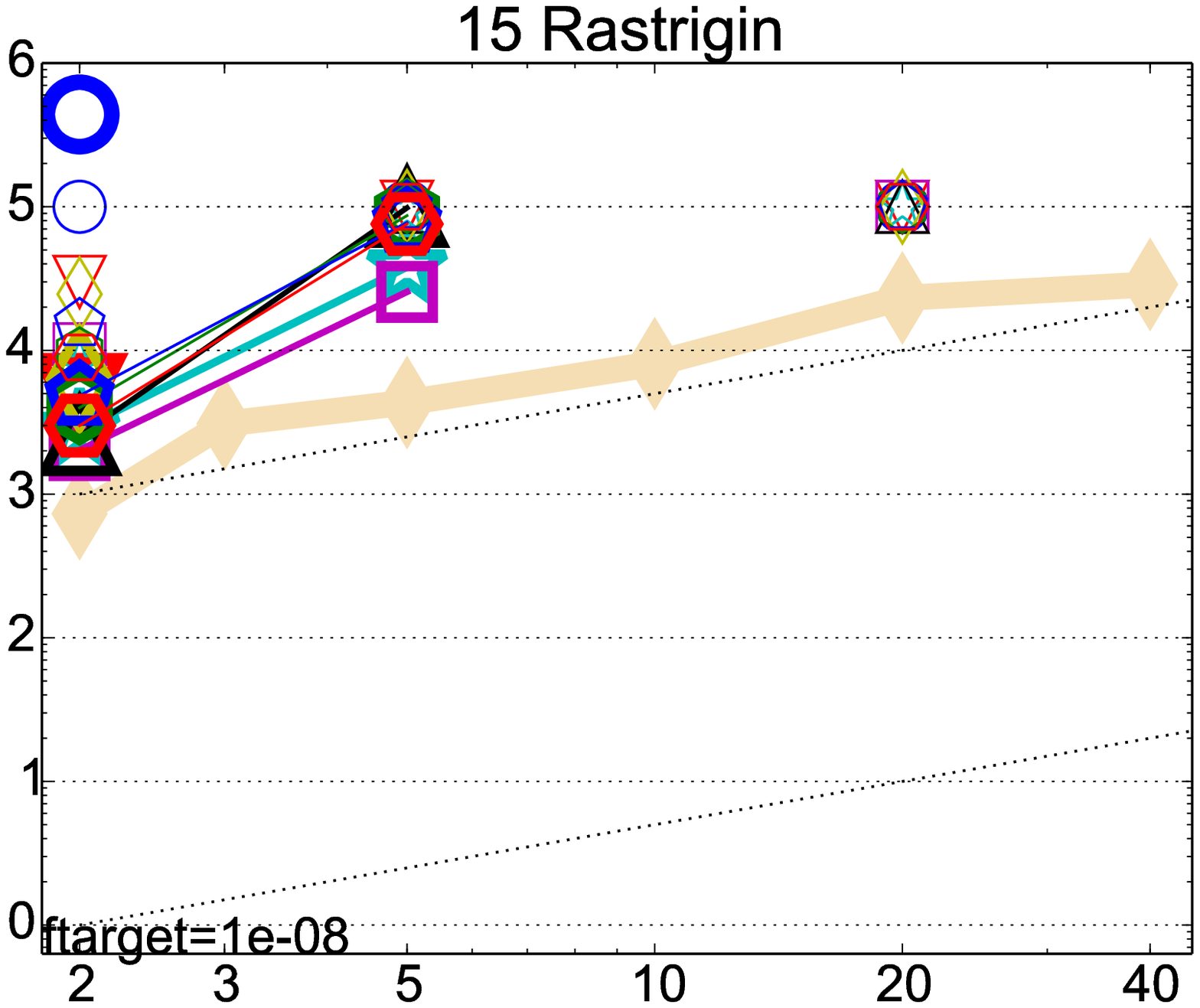}&
\includegraphics[width=0.238\textwidth, trim= 1.8cm 0.8cm 0.5cm 0.5cm, clip]{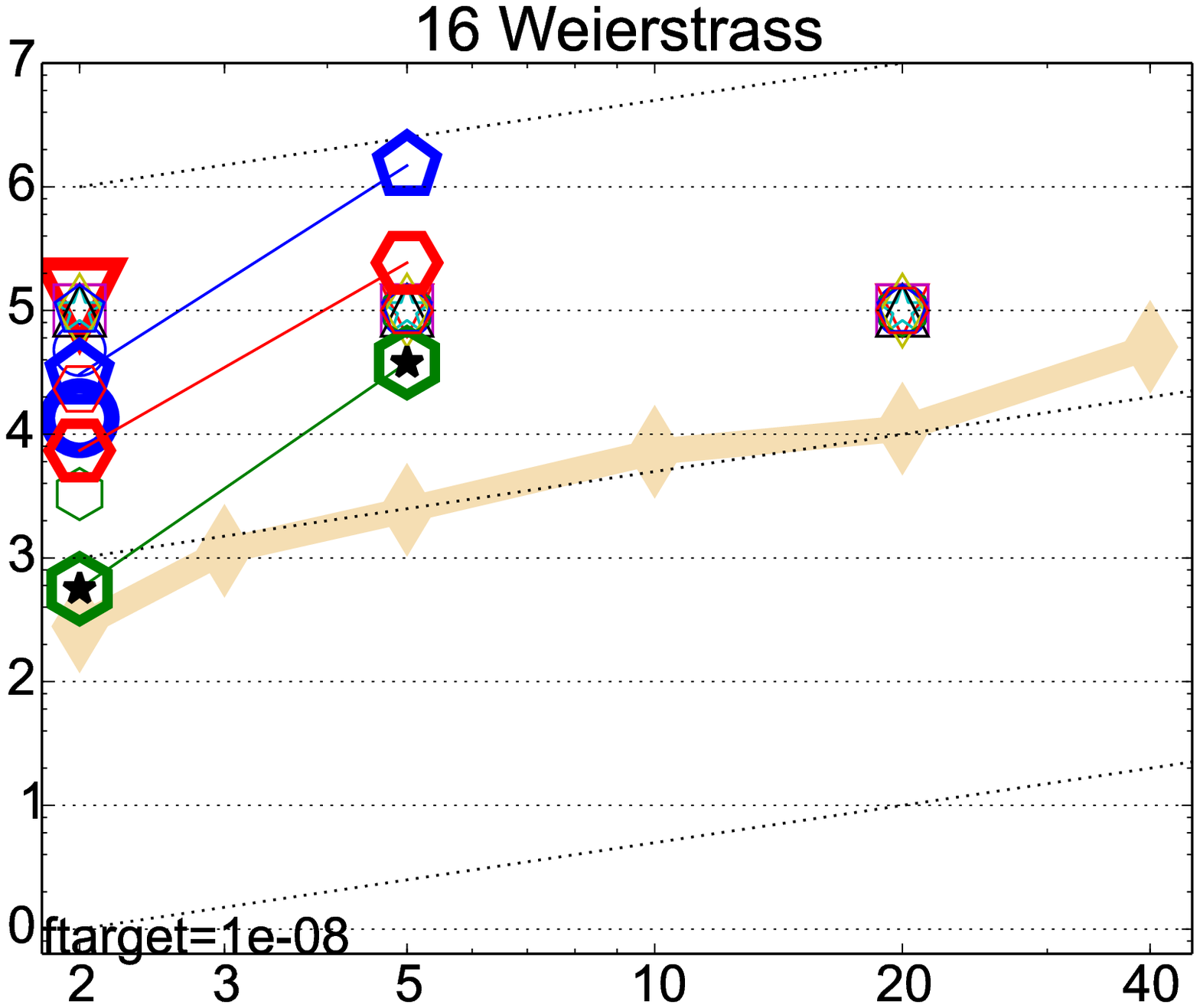}\\
\includegraphics[width=0.253\textwidth, trim= 0.7cm 0.8cm 0.5cm 0.5cm, clip]{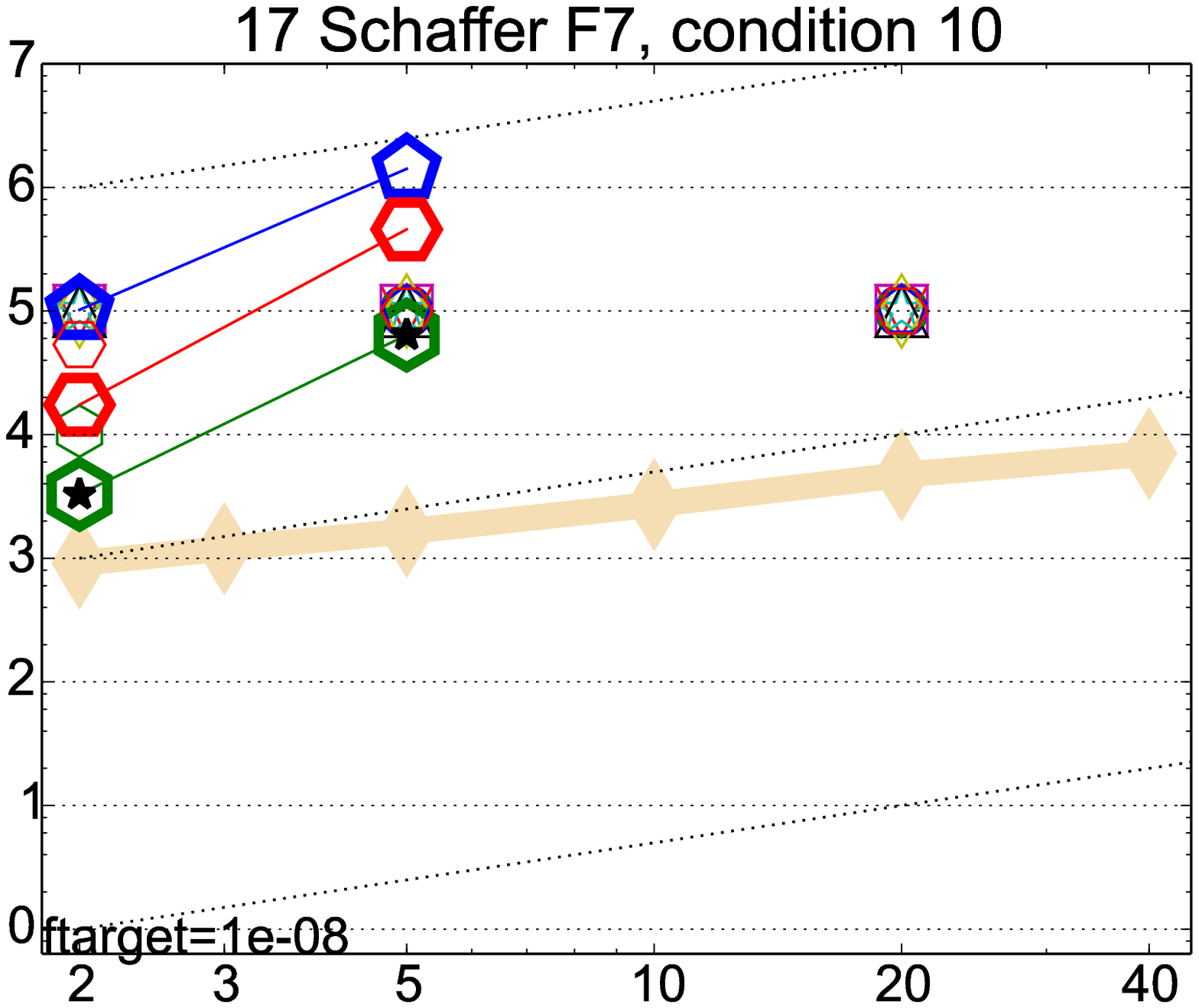}&
\includegraphics[width=0.238\textwidth, trim= 1.8cm 0.8cm 0.5cm 0.5cm, clip]{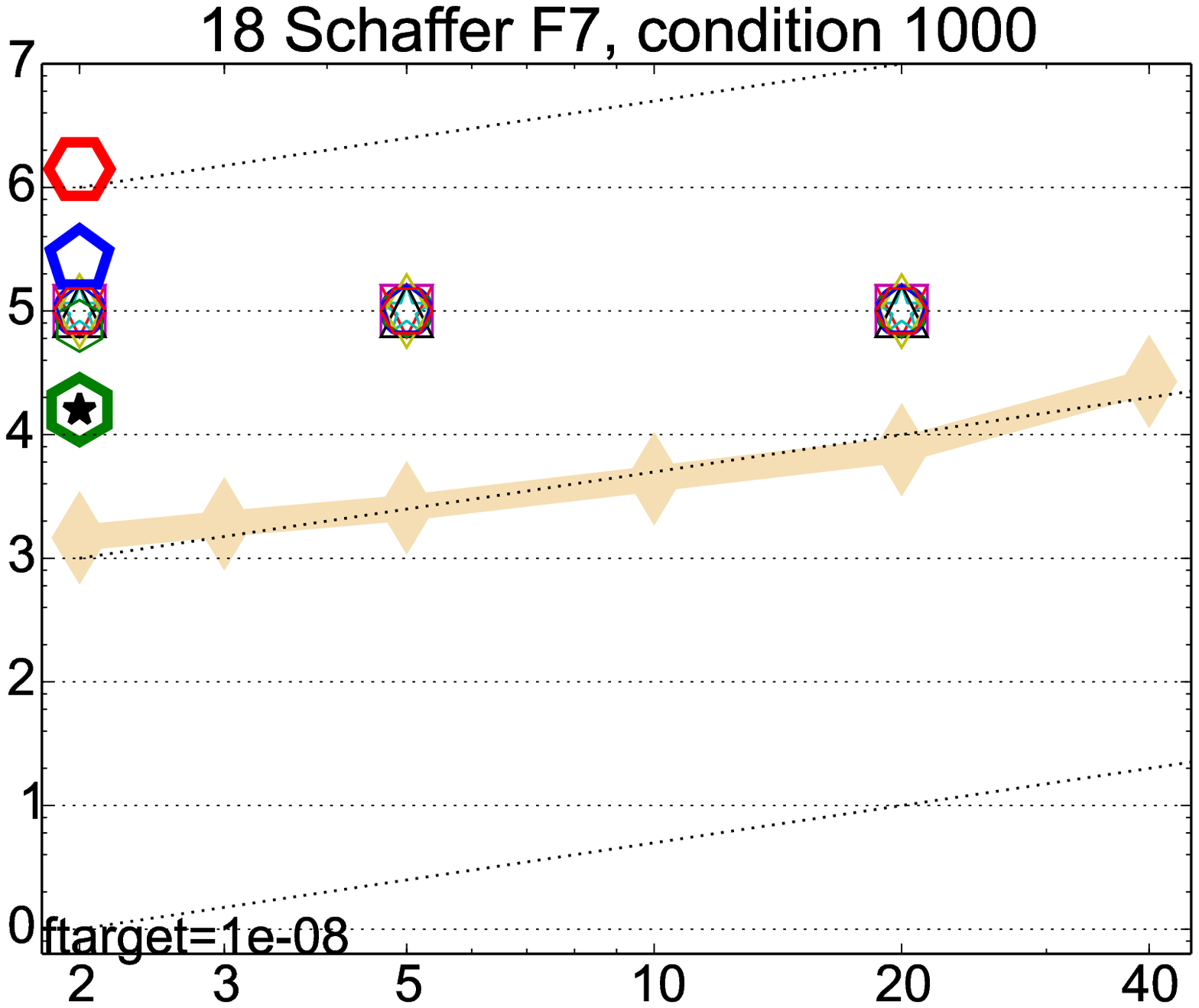}&
\includegraphics[width=0.238\textwidth, trim= 1.8cm 0.8cm 0.5cm 0.5cm, clip]{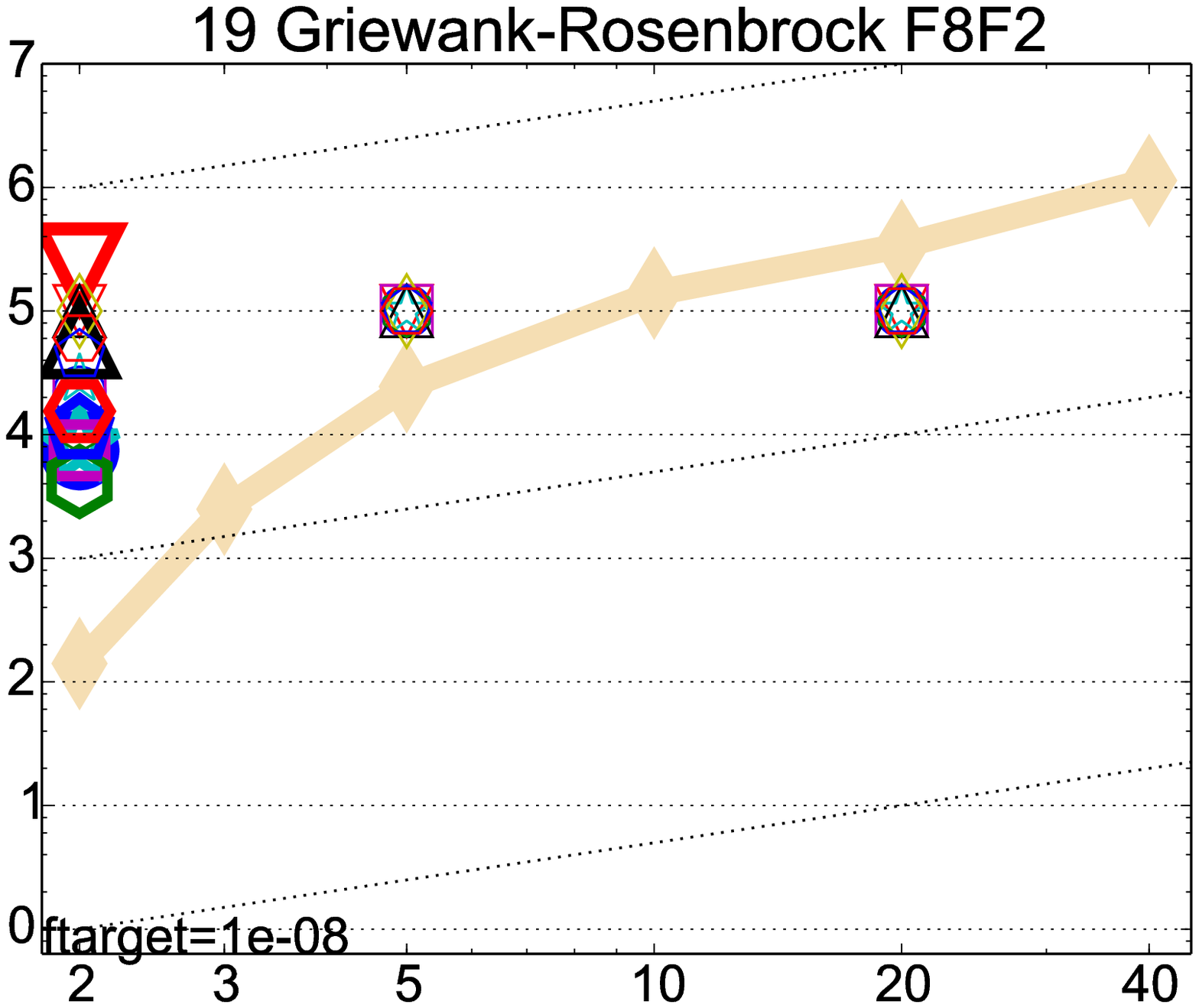}&
\includegraphics[width=0.238\textwidth, trim= 1.8cm 0.8cm 0.5cm 0.5cm, clip]{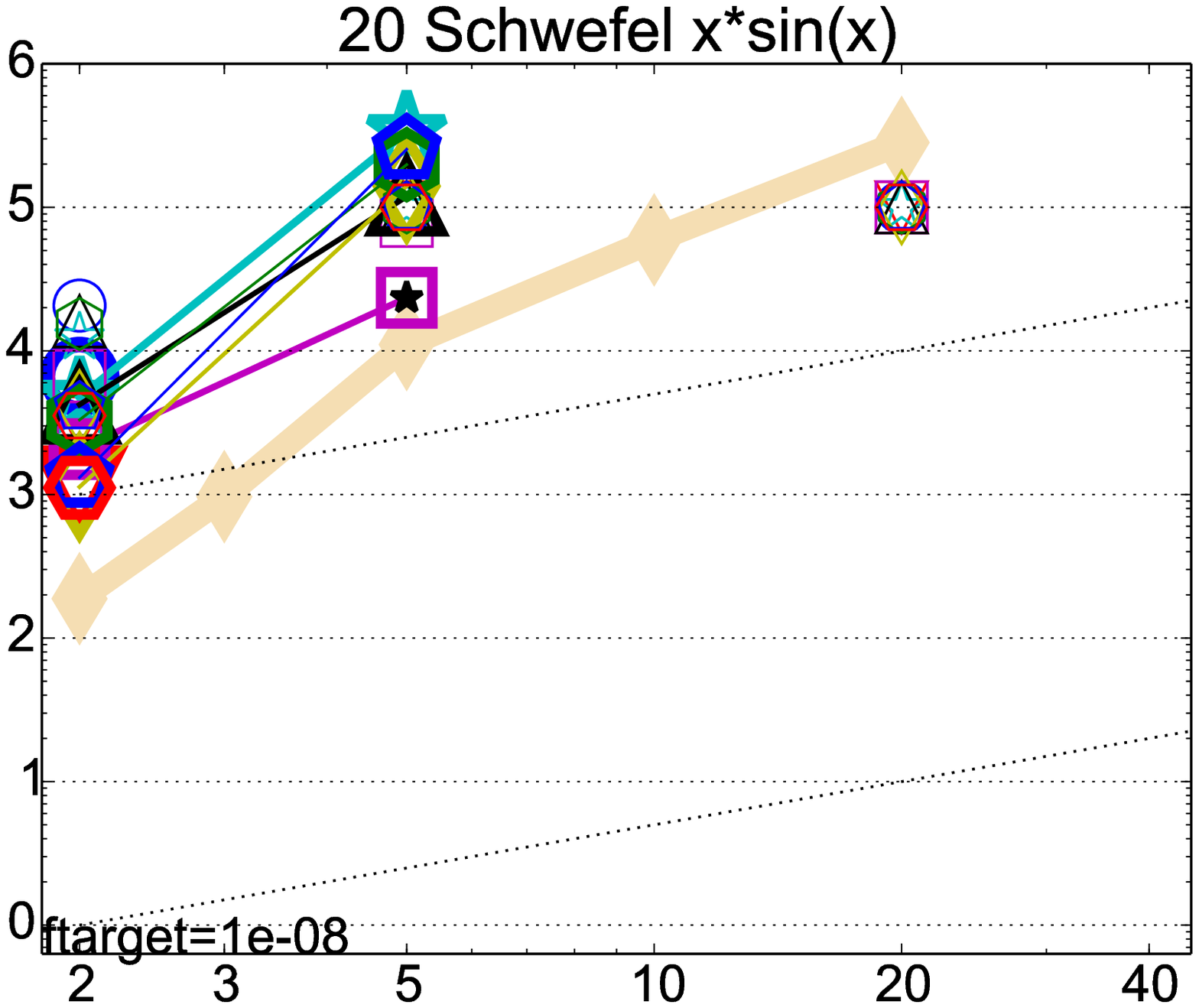}\\
\includegraphics[width=0.253\textwidth, trim= 0.7cm 0.0cm 0.5cm 0.5cm, clip]{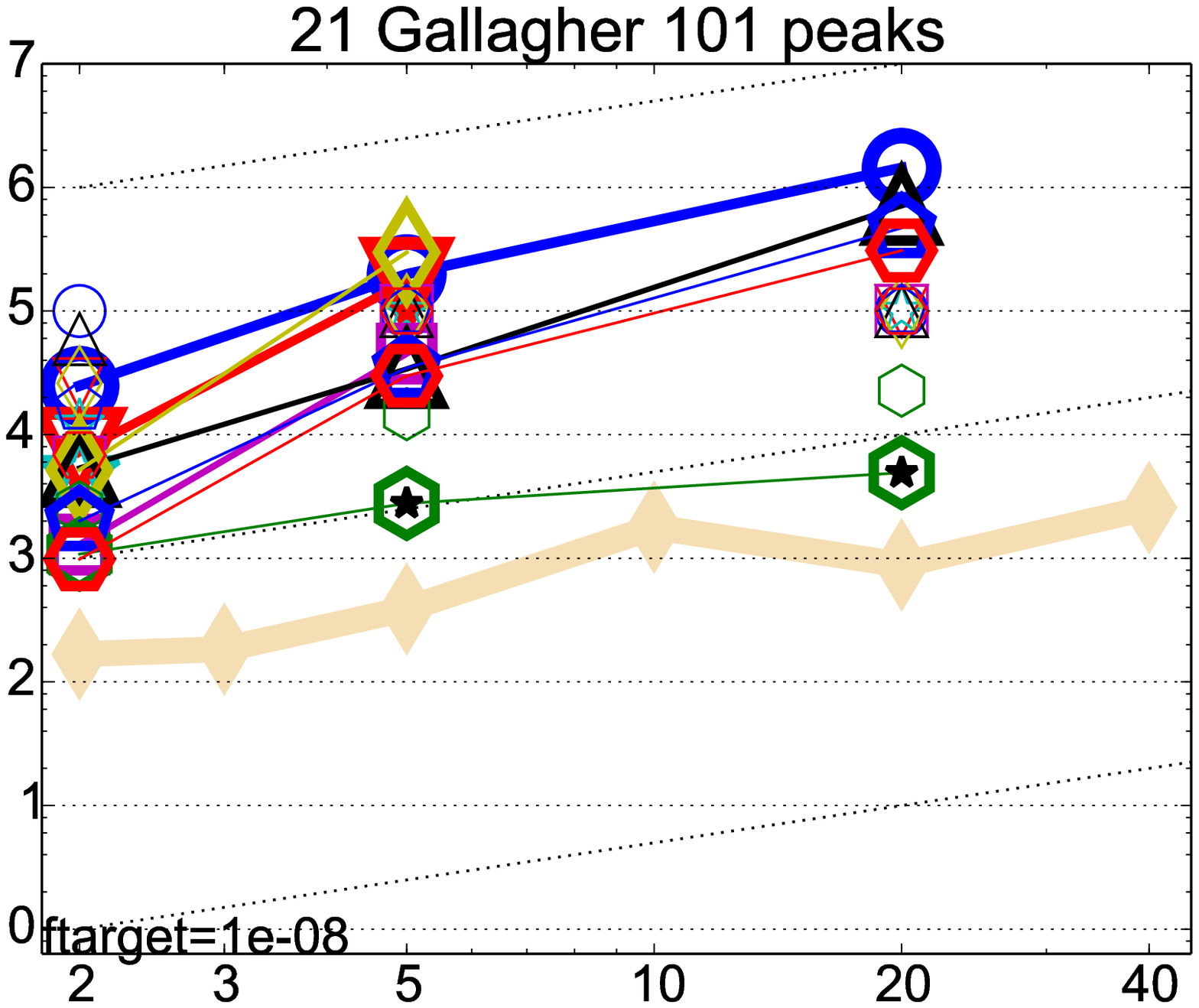}&
\includegraphics[width=0.238\textwidth, trim= 1.8cm 0.0cm 0.5cm 0.5cm, clip]{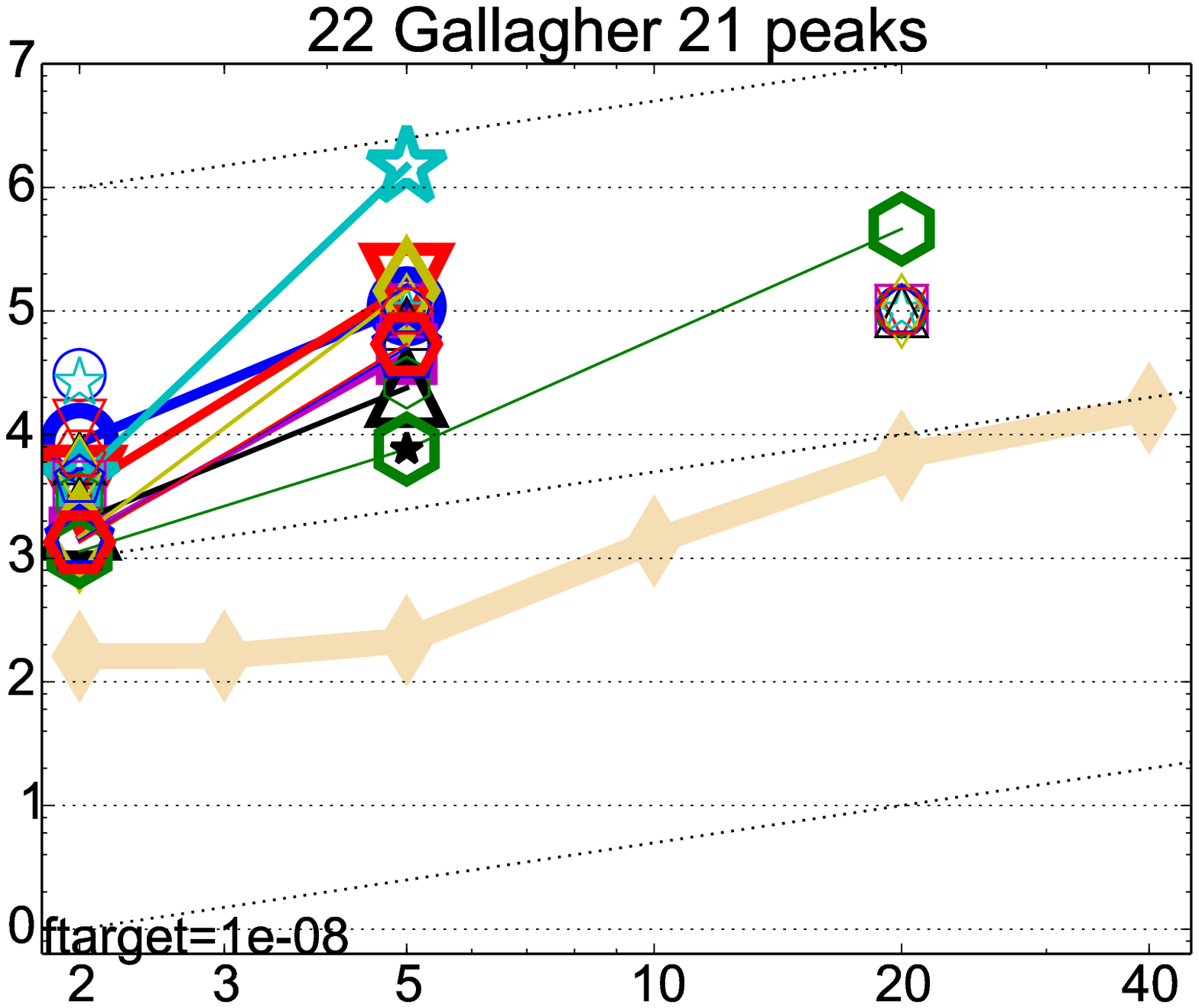}&
\includegraphics[width=0.238\textwidth, trim= 1.8cm 0.0cm 0.5cm 0.5cm, clip]{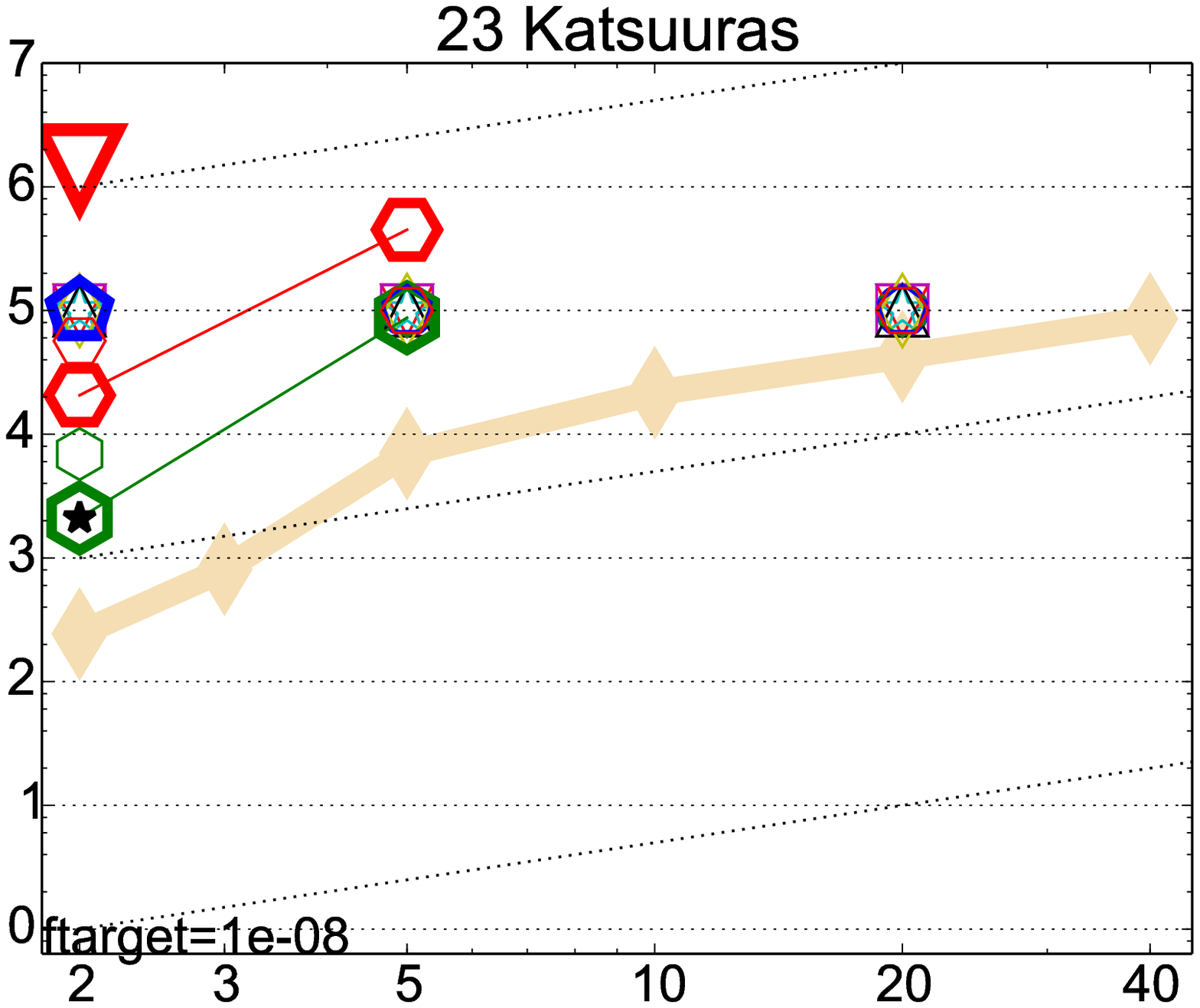}&
\includegraphics[width=0.238\textwidth, trim= 1.8cm 0.0cm 0.5cm 0.5cm, clip]{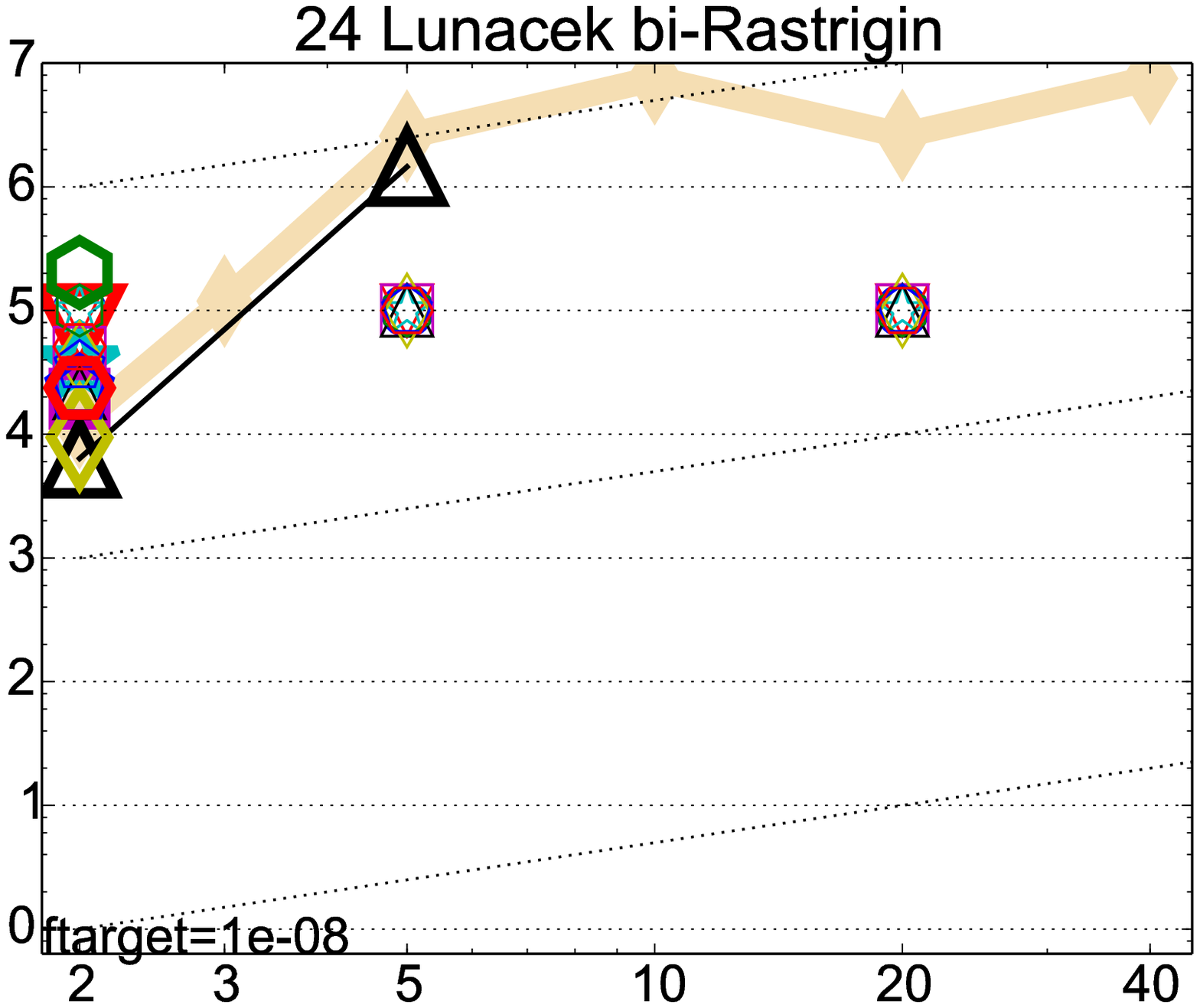}
\end{tabular}
\vspace*{-0.2cm}
\captionwide{Expected running time divided by dimension versus dimension.
\bbobppfigslegend{$f_1$ and $f_{24}$}  
}
\label{figscaling}
\end{figure*}

\newcommand{\rot}[2][2.5]{
  \hspace*{-3.5\baselineskip}%
  \begin{rotate}{90}\hspace{#1em}#2
  \end{rotate}}
\newcommand{\includeperfprof}[1]{
  \input{#1}%
  \includegraphics[width=0.4135\textwidth,trim=0mm 0mm 34mm 10mm, clip]{#1}%
  \raisebox{.037\textwidth}{\parbox[b][.3\textwidth]{.0868\textwidth}{\begin{scriptsize}
    \perfprofsidepanel 
  \end{scriptsize}}}
}
\begin{figure*}
\begin{tabular}{@{}c@{}c@{}}
 separable fcts & moderate fcts \\
  \input{pprldmany_05D_separ}%
  \includegraphics[width=0.4135\textwidth,trim=0mm 0mm 34mm 10mm, clip]{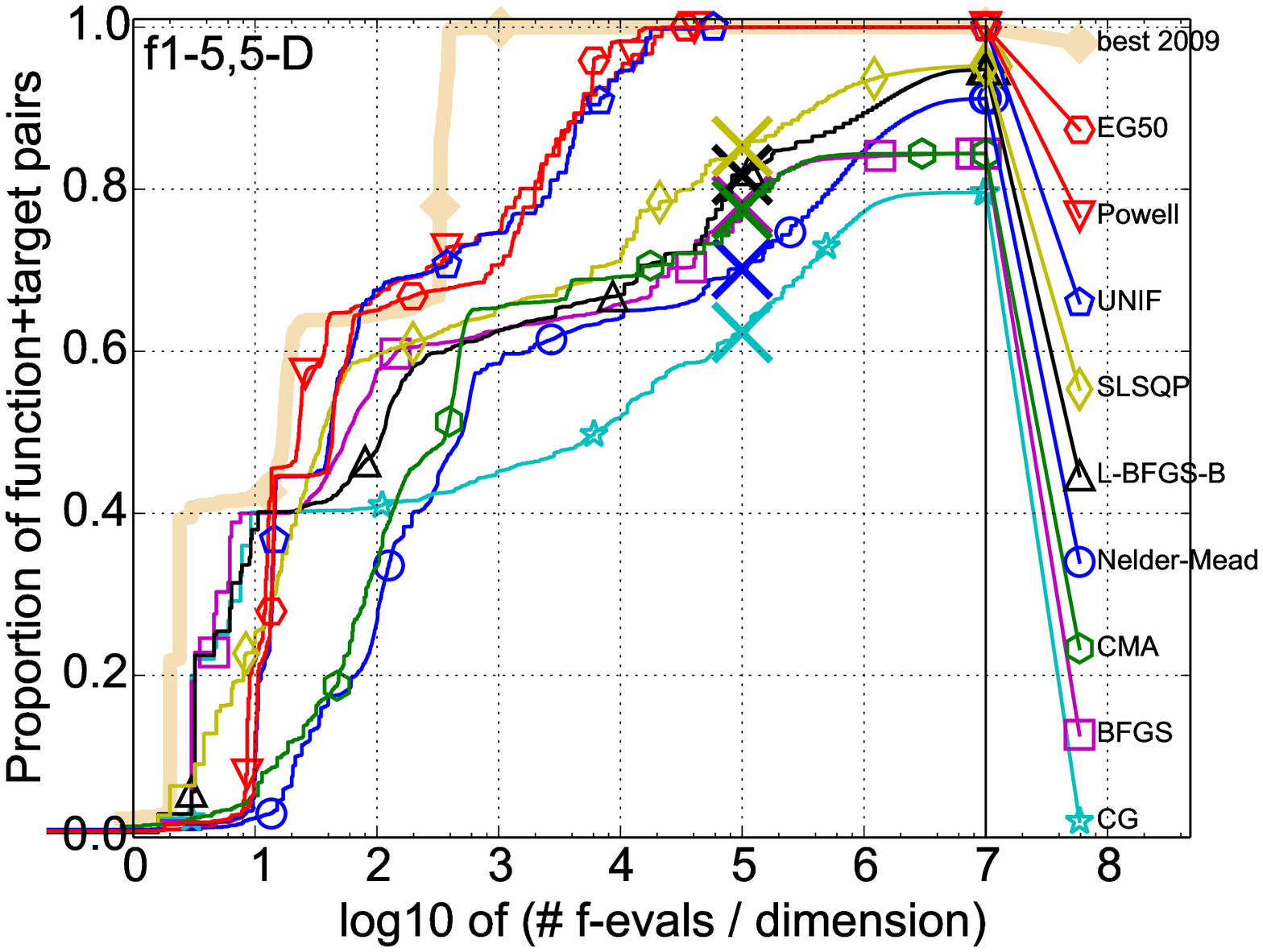}%
  \raisebox{.037\textwidth}{\parbox[b][.3\textwidth]{.0868\textwidth}{\begin{scriptsize}
    \perfprofsidepanel 
  \end{scriptsize}}}
 &
  \input{pprldmany_05D_lcond}%
  \includegraphics[width=0.4135\textwidth,trim=0mm 0mm 34mm 10mm, clip]{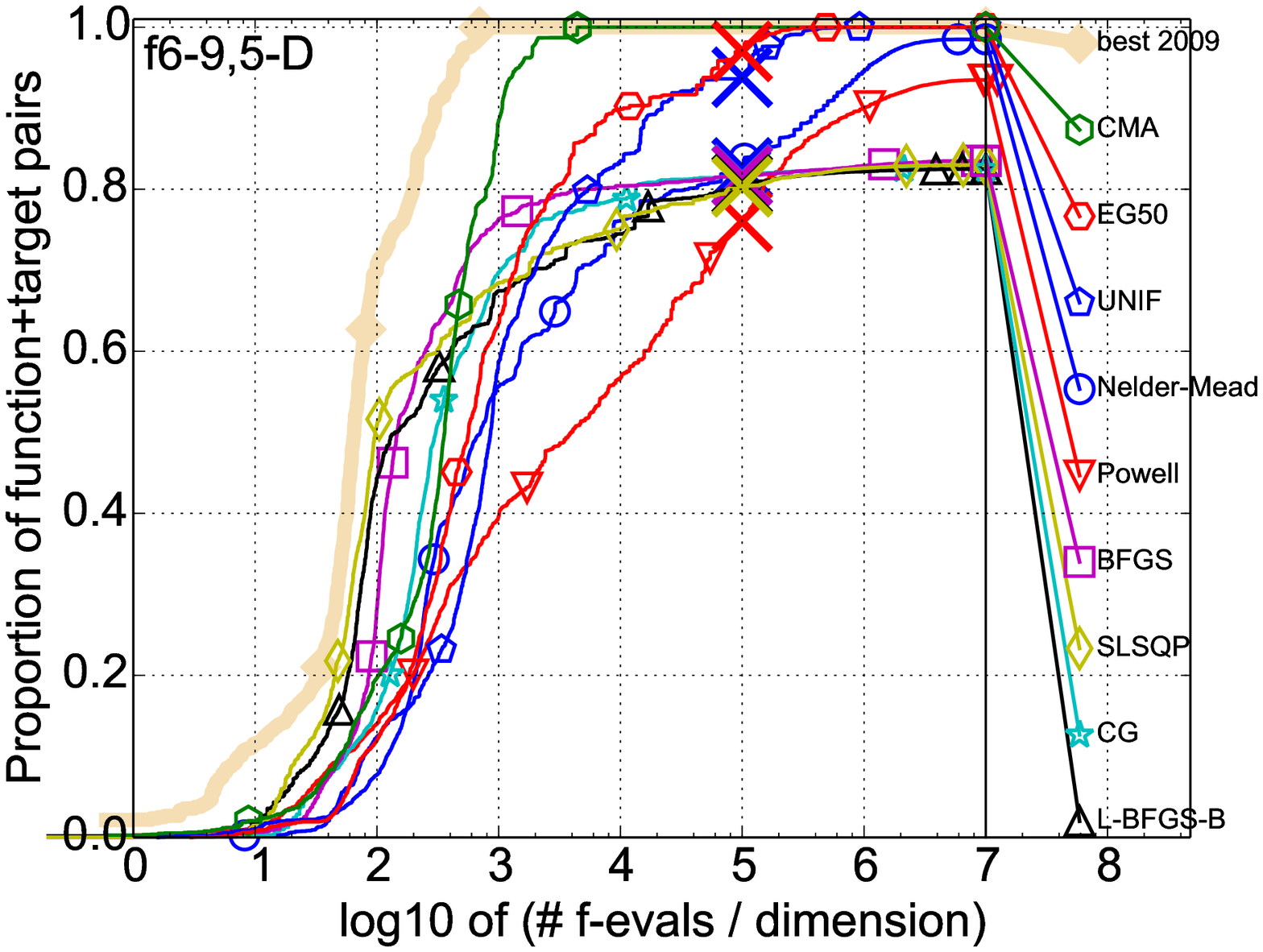}%
  \raisebox{.037\textwidth}{\parbox[b][.3\textwidth]{.0868\textwidth}{\begin{scriptsize}
    \perfprofsidepanel 
  \end{scriptsize}}}
 \\
ill-conditioned fcts & multi-modal fcts \\
  \input{pprldmany_05D_hcond}%
  \includegraphics[width=0.4135\textwidth,trim=0mm 0mm 34mm 10mm, clip]{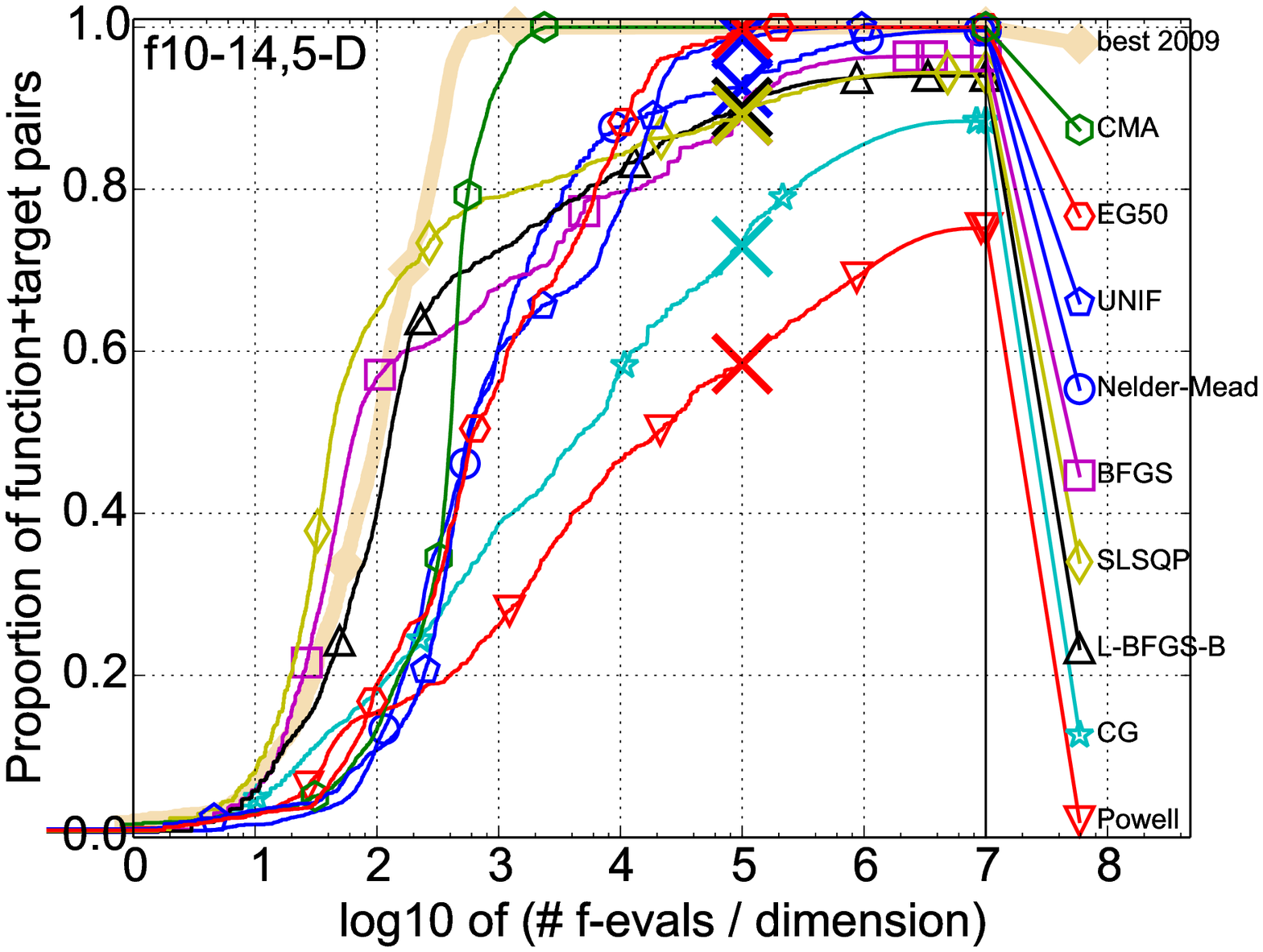}%
  \raisebox{.037\textwidth}{\parbox[b][.3\textwidth]{.0868\textwidth}{\begin{scriptsize}
    \perfprofsidepanel 
  \end{scriptsize}}}
 &
  \input{pprldmany_05D_multi}%
  \includegraphics[width=0.4135\textwidth,trim=0mm 0mm 34mm 10mm, clip]{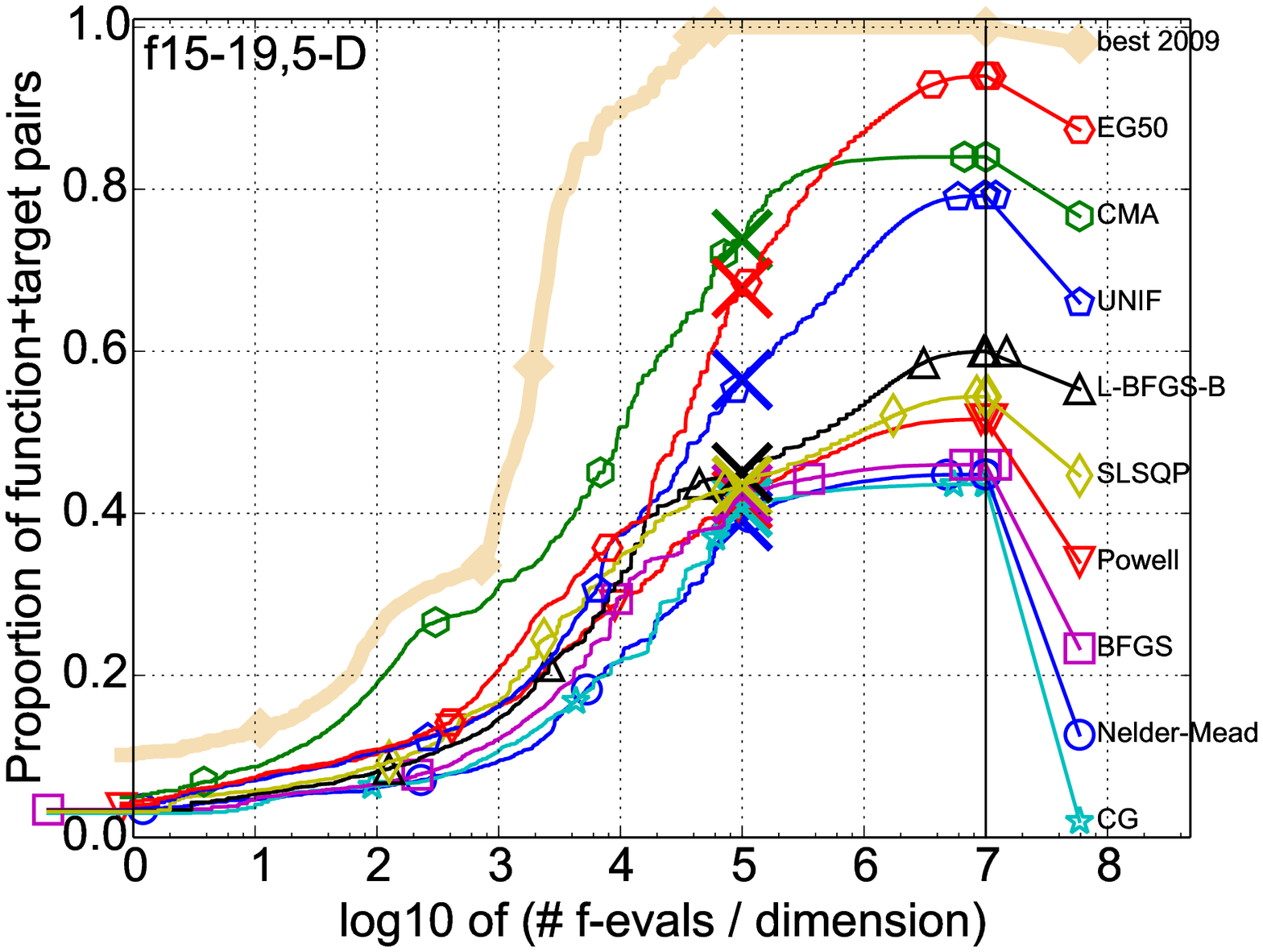}%
  \raisebox{.037\textwidth}{\parbox[b][.3\textwidth]{.0868\textwidth}{\begin{scriptsize}
    \perfprofsidepanel 
  \end{scriptsize}}}
 \\
 weakly structured multi-modal fcts & all functions\\
  \input{pprldmany_05D_mult2}%
  \includegraphics[width=0.4135\textwidth,trim=0mm 0mm 34mm 10mm, clip]{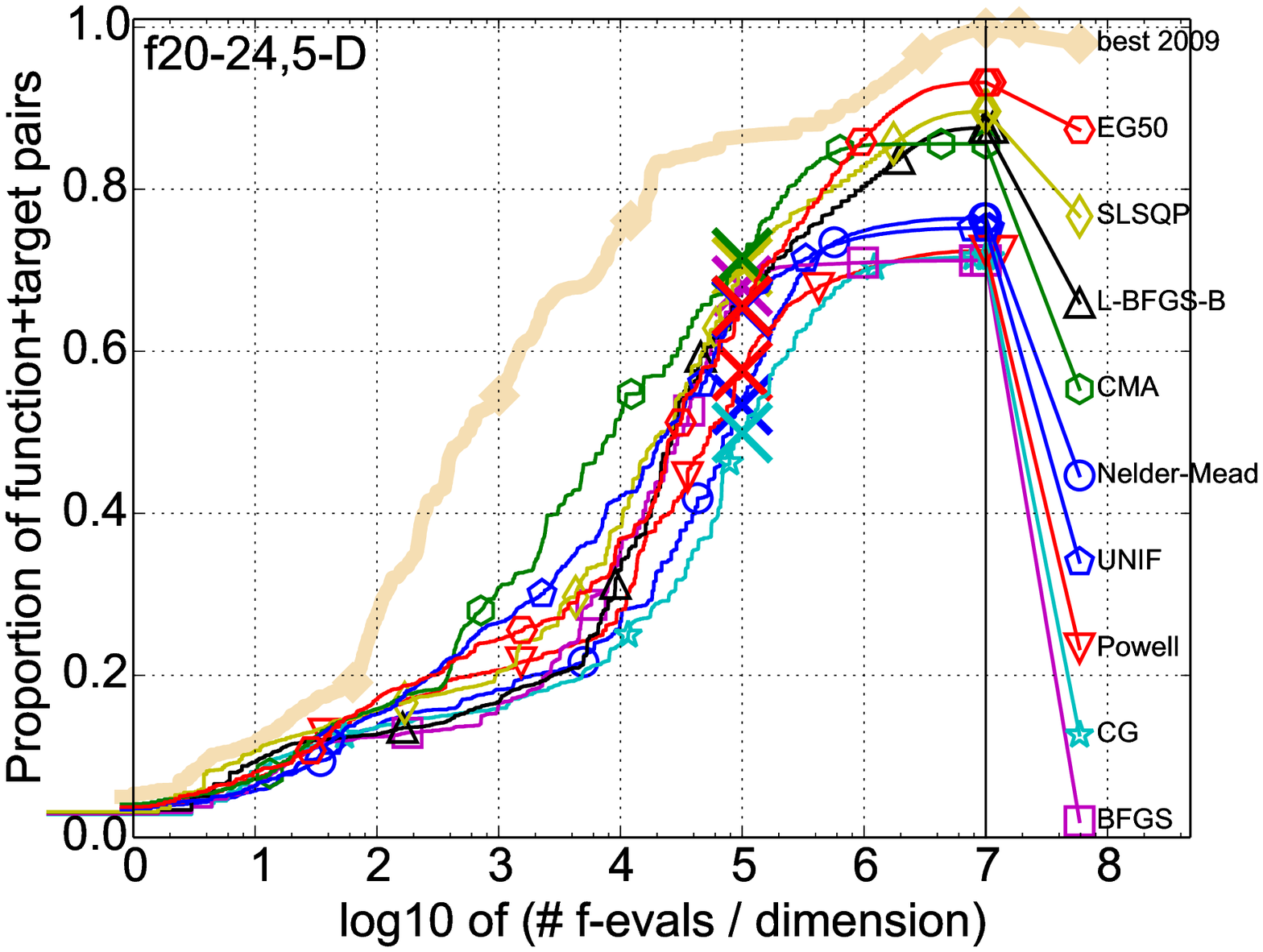}%
  \raisebox{.037\textwidth}{\parbox[b][.3\textwidth]{.0868\textwidth}{\begin{scriptsize}
    \perfprofsidepanel 
  \end{scriptsize}}}
 &
  \input{pprldmany_05D_noiselessall}%
  \includegraphics[width=0.4135\textwidth,trim=0mm 0mm 34mm 10mm, clip]{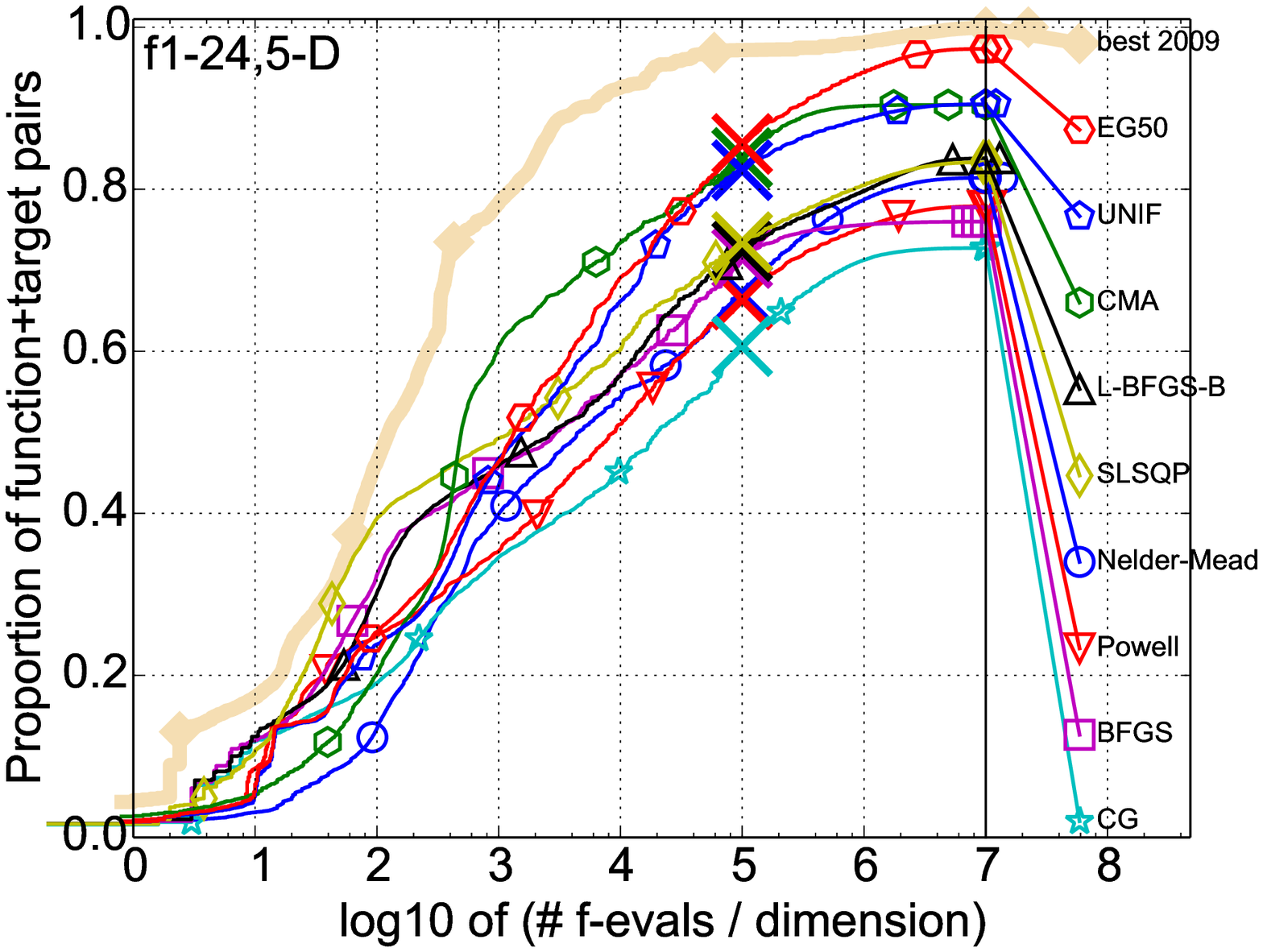}%
  \raisebox{.037\textwidth}{\parbox[b][.3\textwidth]{.0868\textwidth}{\begin{scriptsize}
    \perfprofsidepanel 
  \end{scriptsize}}}

 \end{tabular}
\captionwide{Bootstrapped empirical cumulative distribution of
the number of objective function evaluations
divided by dimension (FEvals/D) for 50 targets in
$10^{[-8..2]}$ for all functions and subgroups in 5-D. The ``best 2009'' line
corresponds to the best \ERT\ observed during BBOB 2009 for each single target.
}
\label{figECDFs05D}
\end{figure*}

\begin{figure*}
 \begin{tabular}{@{}c@{}c@{}}
 separable fcts & moderate fcts \\
  \input{pprldmany_20D_separ}%
  \includegraphics[width=0.4135\textwidth,trim=0mm 0mm 34mm 10mm, clip]{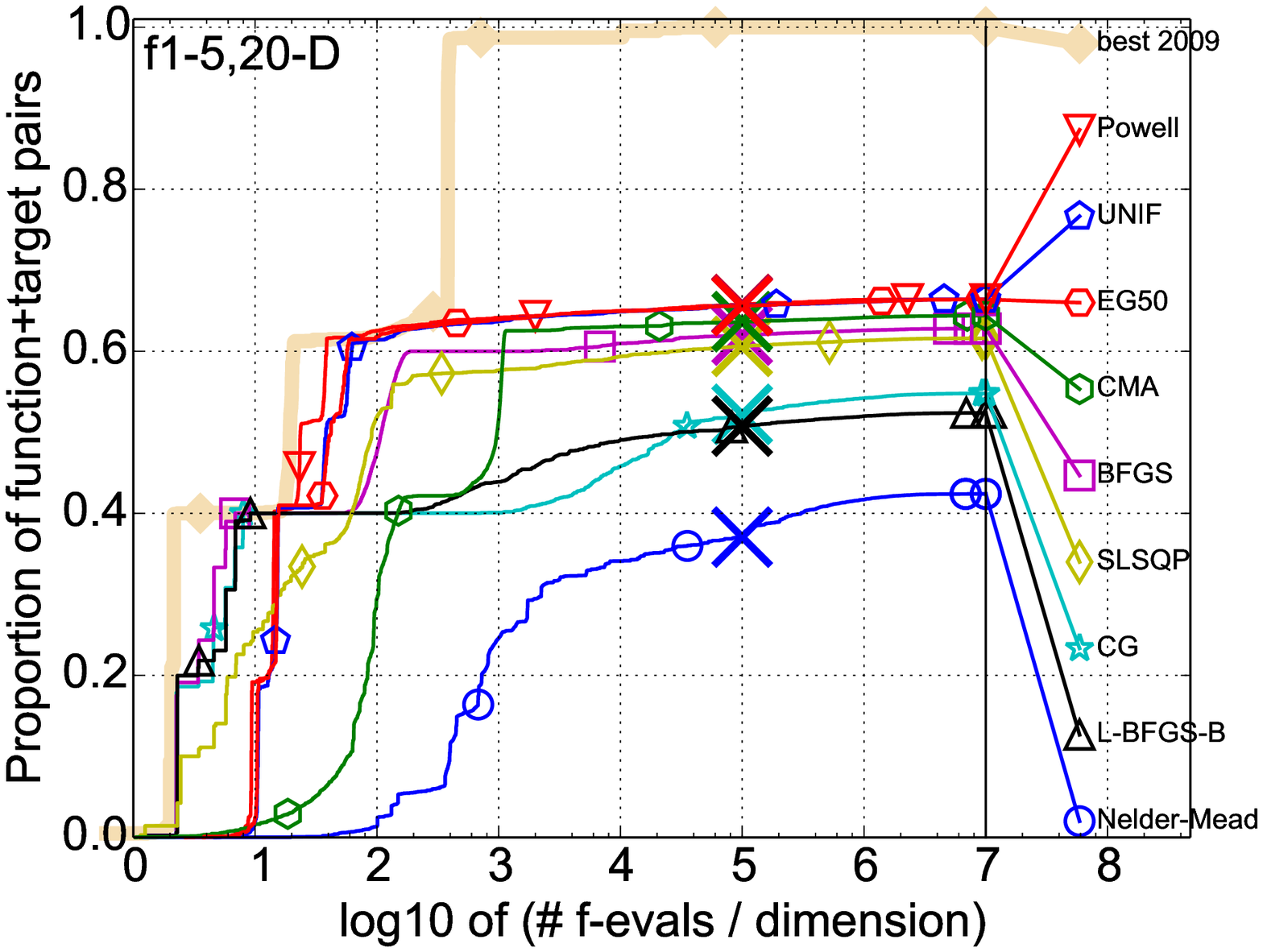}%
  \raisebox{.037\textwidth}{\parbox[b][.3\textwidth]{.0868\textwidth}{\begin{scriptsize}
    \perfprofsidepanel 
  \end{scriptsize}}}
 &
  \input{pprldmany_20D_lcond}%
  \includegraphics[width=0.4135\textwidth,trim=0mm 0mm 34mm 10mm, clip]{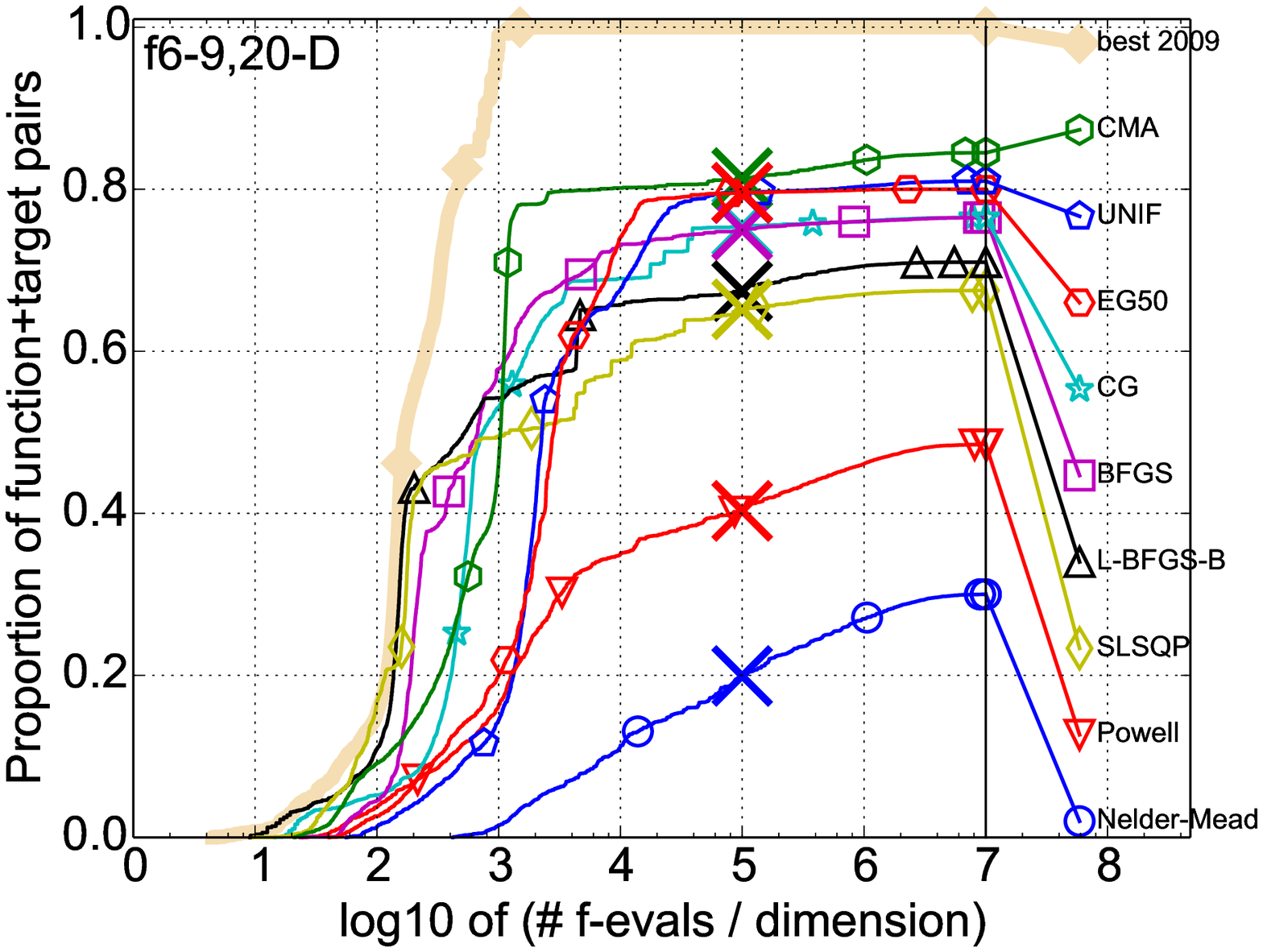}%
  \raisebox{.037\textwidth}{\parbox[b][.3\textwidth]{.0868\textwidth}{\begin{scriptsize}
    \perfprofsidepanel 
  \end{scriptsize}}}
 \\
ill-conditioned fcts & multi-modal fcts \\
  \input{pprldmany_20D_hcond}%
  \includegraphics[width=0.4135\textwidth,trim=0mm 0mm 34mm 10mm, clip]{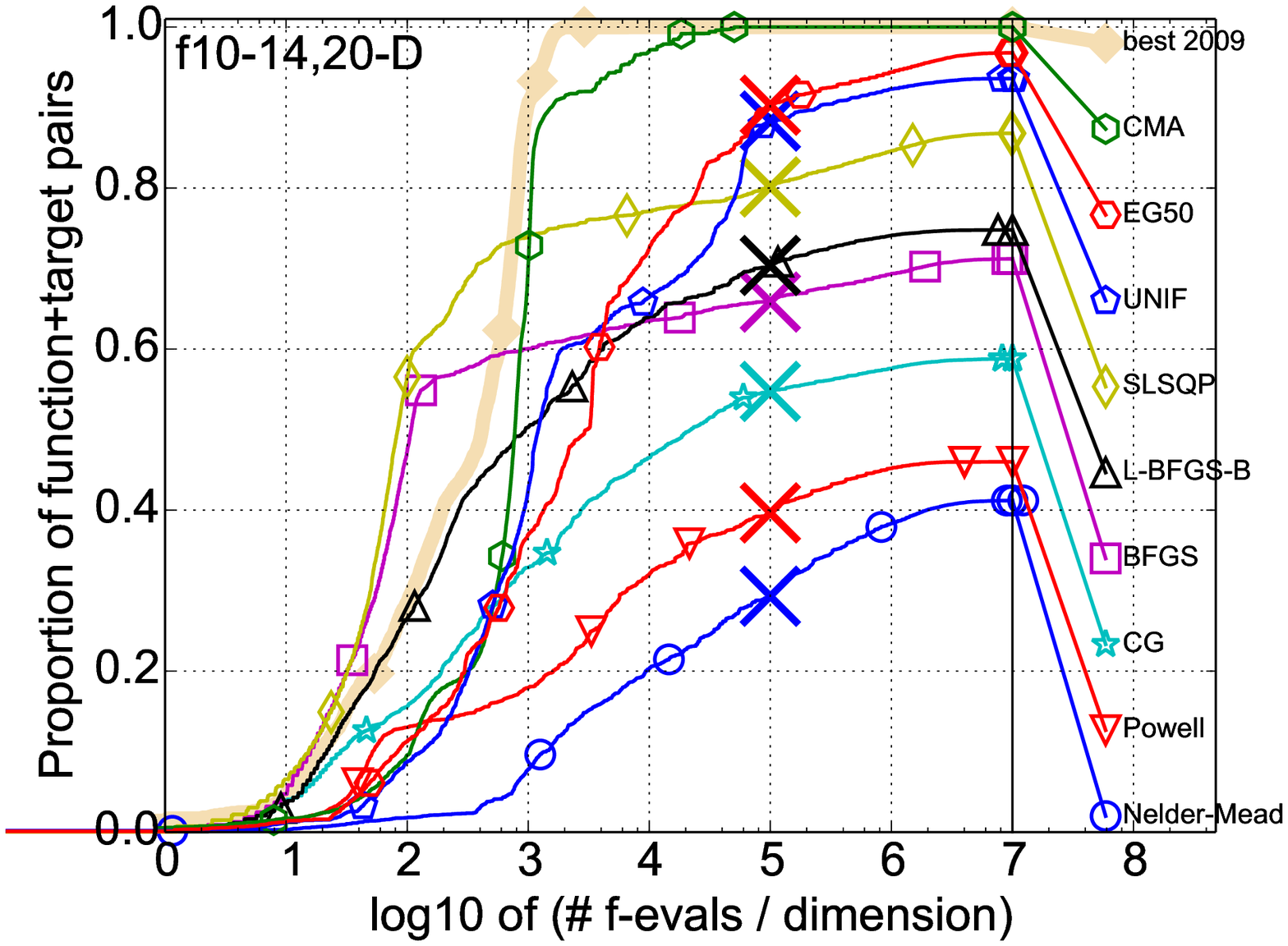}%
  \raisebox{.037\textwidth}{\parbox[b][.3\textwidth]{.0868\textwidth}{\begin{scriptsize}
    \perfprofsidepanel 
  \end{scriptsize}}}
 &
  \input{pprldmany_20D_multi}%
  \includegraphics[width=0.4135\textwidth,trim=0mm 0mm 34mm 10mm, clip]{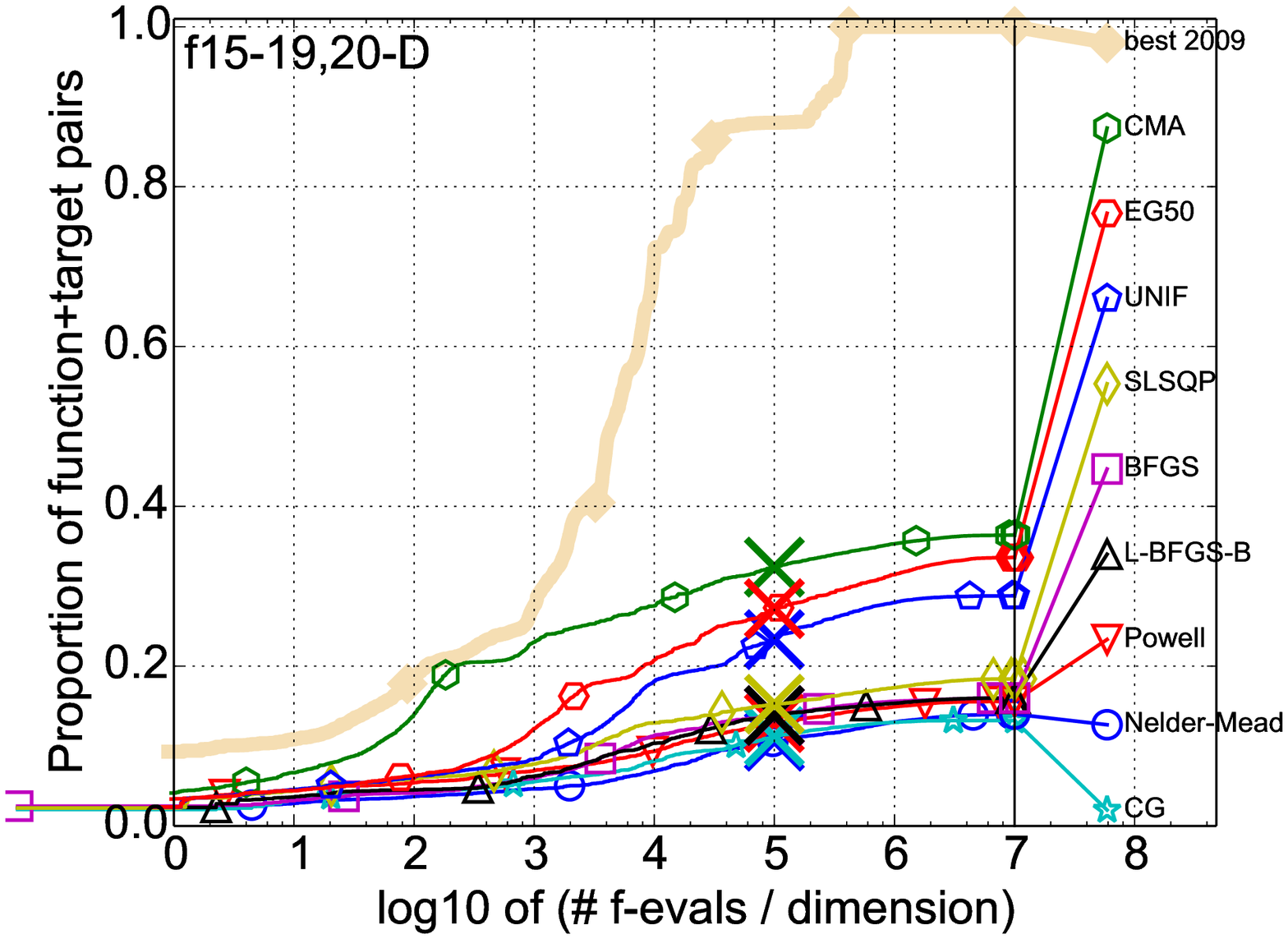}%
  \raisebox{.037\textwidth}{\parbox[b][.3\textwidth]{.0868\textwidth}{\begin{scriptsize}
    \perfprofsidepanel 
  \end{scriptsize}}}
 \\
 weakly structured multi-modal fcts & all functions\\
  \input{pprldmany_20D_mult2}%
  \includegraphics[width=0.4135\textwidth,trim=0mm 0mm 34mm 10mm, clip]{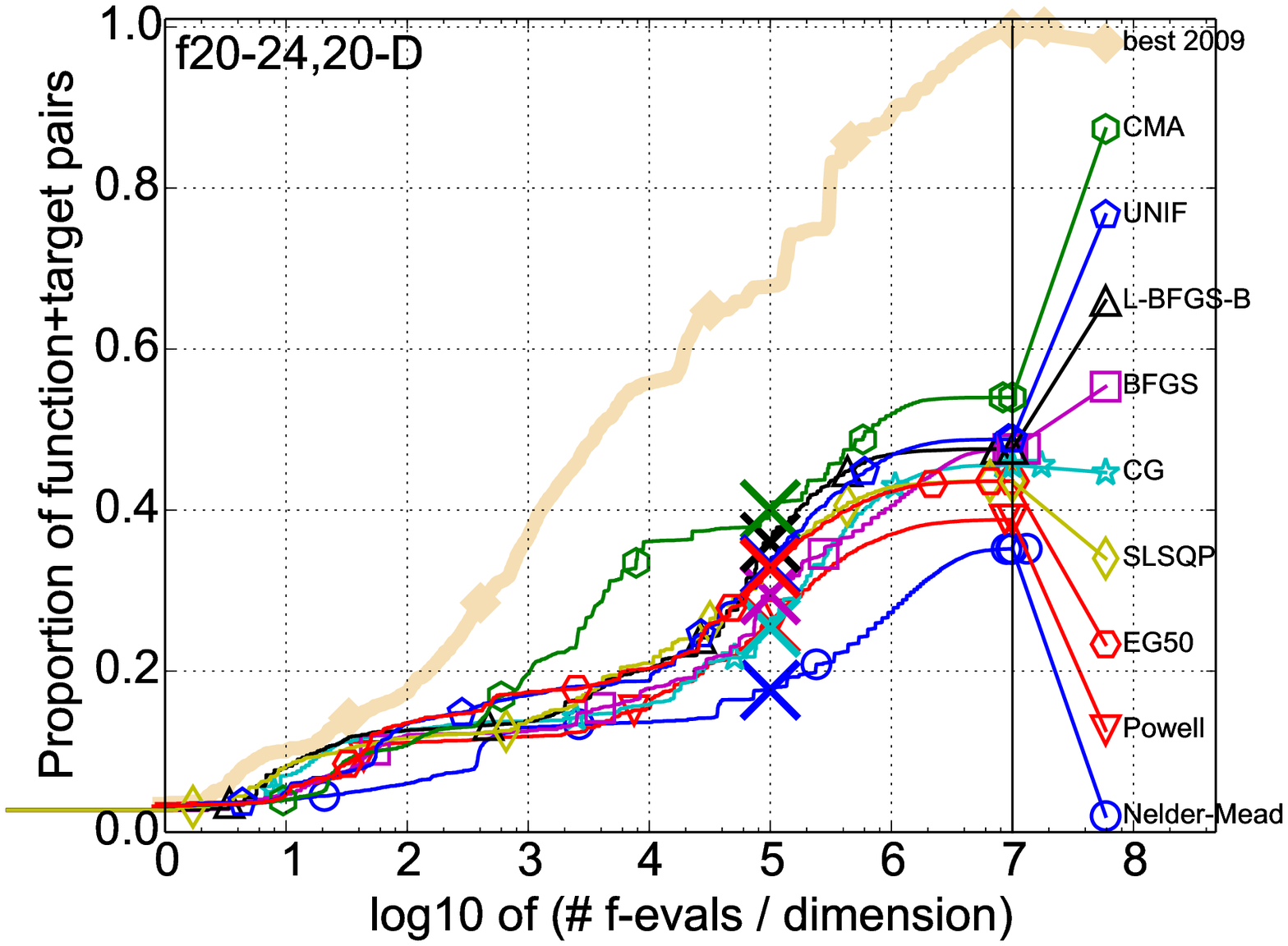}%
  \raisebox{.037\textwidth}{\parbox[b][.3\textwidth]{.0868\textwidth}{\begin{scriptsize}
    \perfprofsidepanel 
  \end{scriptsize}}}
 &
  \input{pprldmany_20D_noiselessall}%
  \includegraphics[width=0.4135\textwidth,trim=0mm 0mm 34mm 10mm, clip]{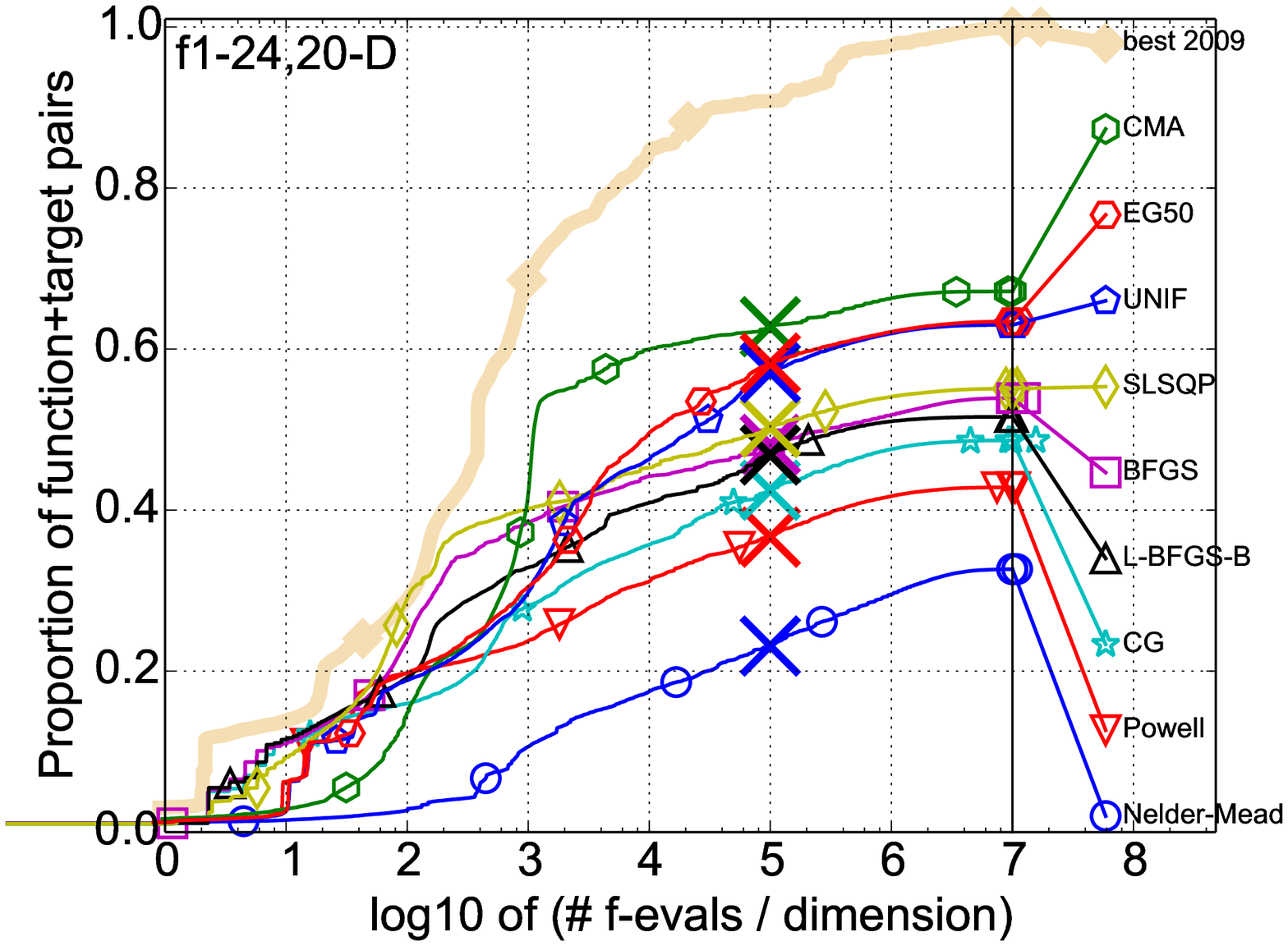}%
  \raisebox{.037\textwidth}{\parbox[b][.3\textwidth]{.0868\textwidth}{\begin{scriptsize}
    \perfprofsidepanel 
  \end{scriptsize}}}

 \end{tabular}
\captionwide{%
Bootstrapped empirical cumulative distribution of
the number of objective function evaluations
divided by dimension (FEvals/D) for 50 targets in
$10^{[-8..2]}$ for all functions and subgroups in 20-D. The ``best 2009'' line
corresponds to the best \ERT\ observed during BBOB 2009 for each single target.
}
\label{figECDFs20D}
\end{figure*}

\section{Discussion and Conclusion}
\label{conclusion}

We described an easy-to-use framework that allows users to plug in
and experiment with custom algorithm selection strategies and then
directly move to use these strategies on practical problems instead
of benchmarks.
The framework currently focuses on online black-box algorithm
portfolios, being careful not to require any modification to stock
and third-party optimization routines besides a simple callback
mechanism that is commonly provided.  The framework could be easily
modified to deal with off-line scenarios too.

Our insights may provide guidance for users of the SciPy optimizers.
Regarding algorithm portfolios, the results show that even a very naive algorithm selection strategy
can be beneficial; while it cannot beat the best algorithm CMA
in high-dimensional cases on average over all functions, it provides a more consistent
performance than CMA, 
translating to best-of-all performance in low dimensions.

\subsection{Future Work}

The currently chosen reference portfolio is somewhat ad hoc,
chosen in part to contribute value to the SciPy project,
but not optimal for further general research into
algorithm portfolios. In the long run, we hope to transition
to a better balanced portfolio.

A common scheme of online algorithm selection schemes is that
a {\em credit} is {\em assigned} to an algorithm in each round based
on its performance (e.g. relative improvement, absolute value)
and {\em accrued} over the rounds (e.g. averaging it or using
an adaptation rule).  Credit-based selection is so common
that it should be directly supported by the framework; diverse
pluggable strategies would then enable exploration
of various interesting strategy combinations.  A work-in-progress
preview implementation is already available.

Actually {\em running} the individual
algorithms is appropriate for optimizing non-benchmark
functions and useful in the future when
solutions may migrate within the portfolio population. However,
a~useful feature would be to allow testing selection strategies based on just
the saved ``mlog`` data (see Sec. \ref{mlog}) from individual algorithm runs.


%
%
%
%

\section*{Acknowledgements}
Research described in the paper was supervised by Dr. Petr Pošík 
and supported by the CTU grant SGS14/194/ OHK3/3T/13 ``Automatic adaptation of search algorithms''.




\bibliographystyle{acm}
\bibliography{cocopf,bbob}

\begin{authorcv}{Petr Baudiš}
	has received his Bachelors degree and Masters degree in
	Theoretical Computer Science at the Charles University in Prague.
	His Master thesis ``MCTS with Information Sharing'' presented a state-of-art
	Computer Go program. Currently, he is a PhD student at the Czech Technical University
	with principal research interest in Algorithm Portfolios
	and applications of Monte Carlo Tree Search.
\end{authorcv}

\end{document}

%% file: bbob_pproc_commands.tex
\providecommand{\bbobppfigsftarget}{\ensuremath{10^{-8}}}
\providecommand{\bbobppfigslegend}[1]{
Expected running time (\ERT\ in number of $f$-evaluations) divided by dimension for target function value \bbobppfigsftarget\ as $\log_{10}$ values versus dimension. Light symbols give the maximum number of function evaluations from the longest trial divided by dimension. Horizontal lines give linear scaling, slanted dotted lines give quadratic scaling. Black stars indicate statistically better result compared to all other algorithms with $p<0.01$ and Bonferroni correction number of dimensions (six).  
Legend: 
{\color{blue}$\circ$}:\algorithmA
, {\color{red}$\triangledown$}:\algorithmB
, {\color{cyan}$\star$}:\algorithmC
, {\color{magenta}$\Box$}:\algorithmD
, {\color{black}$\triangle$}:\algorithmE
, {\color{yellow}$\diamondsuit$}:\algorithmF
, {\color{green!45!black}$\varhexagon$}:\algorithmG
, {\color{blue}$\pentagon$}:\algorithmH
, {\color{red}$\hexagon$}:\algorithmI
}

\providecommand{\algorithmA}{Nelder-Mead}
\providecommand{\algorithmB}{Powell}
\providecommand{\algorithmC}{CG}
\providecommand{\algorithmD}{BFGS}
\providecommand{\algorithmE}{L-BFGS-B}
\providecommand{\algorithmF}{SLSQP}
\providecommand{\algorithmG}{CMA}
\providecommand{\algorithmH}{UNIF}
\providecommand{\algorithmI}{EG50}
\providecommand{\algorithmA}{Nelder-Mead}
\providecommand{\algorithmB}{Powell}
\providecommand{\algorithmC}{CG}
\providecommand{\algorithmD}{BFGS}
\providecommand{\algorithmE}{L-BFGS-B}
\providecommand{\algorithmF}{SLSQP}
\providecommand{\algorithmG}{CMA}
\providecommand{\algorithmH}{UNIF}
\providecommand{\algorithmI}{EG50}

%% file: pprldmany_05D_separ.tex
\providecommand{\nperfprof}{7}\providecommand{\algaperfprof}{\StrLeft{Nelder-Mead}{\nperfprof}}\providecommand{\algbperfprof}{\StrLeft{Powell}{\nperfprof}}\providecommand{\algcperfprof}{\StrLeft{CG}{\nperfprof}}\providecommand{\algdperfprof}{\StrLeft{BFGS}{\nperfprof}}\providecommand{\algeperfprof}{\StrLeft{L-BFGS-B}{\nperfprof}}\providecommand{\algfperfprof}{\StrLeft{SLSQP}{\nperfprof}}\providecommand{\alggperfprof}{\StrLeft{CMA}{\nperfprof}}\providecommand{\alghperfprof}{\StrLeft{UNIF}{\nperfprof}}\providecommand{\algiperfprof}{\StrLeft{EG50}{\nperfprof}}\providecommand{\algzeroperfprof}{best 2009}\providecommand{\perfprofsidepanel}{\mbox{\algzeroperfprof}
\vfill \mbox{\algiperfprof}
\vfill \mbox{\algbperfprof}
\vfill \mbox{\alghperfprof}
\vfill \mbox{\algfperfprof}
\vfill \mbox{\algeperfprof}
\vfill \mbox{\algaperfprof}
\vfill \mbox{\alggperfprof}
\vfill \mbox{\algdperfprof}
\vfill \mbox{\algcperfprof}}

%% file: pprldmany_05D_lcond.tex
\providecommand{\nperfprof}{7}\providecommand{\algaperfprof}{\StrLeft{Nelder-Mead}{\nperfprof}}\providecommand{\algbperfprof}{\StrLeft{Powell}{\nperfprof}}\providecommand{\algcperfprof}{\StrLeft{CG}{\nperfprof}}\providecommand{\algdperfprof}{\StrLeft{BFGS}{\nperfprof}}\providecommand{\algeperfprof}{\StrLeft{L-BFGS-B}{\nperfprof}}\providecommand{\algfperfprof}{\StrLeft{SLSQP}{\nperfprof}}\providecommand{\alggperfprof}{\StrLeft{CMA}{\nperfprof}}\providecommand{\alghperfprof}{\StrLeft{UNIF}{\nperfprof}}\providecommand{\algiperfprof}{\StrLeft{EG50}{\nperfprof}}\providecommand{\algzeroperfprof}{best 2009}\providecommand{\perfprofsidepanel}{\mbox{\algzeroperfprof}
\vfill \mbox{\alggperfprof}
\vfill \mbox{\algiperfprof}
\vfill \mbox{\alghperfprof}
\vfill \mbox{\algaperfprof}
\vfill \mbox{\algbperfprof}
\vfill \mbox{\algdperfprof}
\vfill \mbox{\algfperfprof}
\vfill \mbox{\algcperfprof}
\vfill \mbox{\algeperfprof}}

%% file: pprldmany_05D_hcond.tex
\providecommand{\nperfprof}{7}\providecommand{\algaperfprof}{\StrLeft{Nelder-Mead}{\nperfprof}}\providecommand{\algbperfprof}{\StrLeft{Powell}{\nperfprof}}\providecommand{\algcperfprof}{\StrLeft{CG}{\nperfprof}}\providecommand{\algdperfprof}{\StrLeft{BFGS}{\nperfprof}}\providecommand{\algeperfprof}{\StrLeft{L-BFGS-B}{\nperfprof}}\providecommand{\algfperfprof}{\StrLeft{SLSQP}{\nperfprof}}\providecommand{\alggperfprof}{\StrLeft{CMA}{\nperfprof}}\providecommand{\alghperfprof}{\StrLeft{UNIF}{\nperfprof}}\providecommand{\algiperfprof}{\StrLeft{EG50}{\nperfprof}}\providecommand{\algzeroperfprof}{best 2009}\providecommand{\perfprofsidepanel}{\mbox{\algzeroperfprof}
\vfill \mbox{\alggperfprof}
\vfill \mbox{\algiperfprof}
\vfill \mbox{\alghperfprof}
\vfill \mbox{\algaperfprof}
\vfill \mbox{\algdperfprof}
\vfill \mbox{\algfperfprof}
\vfill \mbox{\algeperfprof}
\vfill \mbox{\algcperfprof}
\vfill \mbox{\algbperfprof}}

%% file: pprldmany_05D_multi.tex
\providecommand{\nperfprof}{7}\providecommand{\algaperfprof}{\StrLeft{Nelder-Mead}{\nperfprof}}\providecommand{\algbperfprof}{\StrLeft{Powell}{\nperfprof}}\providecommand{\algcperfprof}{\StrLeft{CG}{\nperfprof}}\providecommand{\algdperfprof}{\StrLeft{BFGS}{\nperfprof}}\providecommand{\algeperfprof}{\StrLeft{L-BFGS-B}{\nperfprof}}\providecommand{\algfperfprof}{\StrLeft{SLSQP}{\nperfprof}}\providecommand{\alggperfprof}{\StrLeft{CMA}{\nperfprof}}\providecommand{\alghperfprof}{\StrLeft{UNIF}{\nperfprof}}\providecommand{\algiperfprof}{\StrLeft{EG50}{\nperfprof}}\providecommand{\algzeroperfprof}{best 2009}\providecommand{\perfprofsidepanel}{\mbox{\algzeroperfprof}
\vfill \mbox{\algiperfprof}
\vfill \mbox{\alggperfprof}
\vfill \mbox{\alghperfprof}
\vfill \mbox{\algeperfprof}
\vfill \mbox{\algfperfprof}
\vfill \mbox{\algbperfprof}
\vfill \mbox{\algdperfprof}
\vfill \mbox{\algaperfprof}
\vfill \mbox{\algcperfprof}}

%% file: pprldmany_05D_mult2.tex
\providecommand{\nperfprof}{7}\providecommand{\algaperfprof}{\StrLeft{Nelder-Mead}{\nperfprof}}\providecommand{\algbperfprof}{\StrLeft{Powell}{\nperfprof}}\providecommand{\algcperfprof}{\StrLeft{CG}{\nperfprof}}\providecommand{\algdperfprof}{\StrLeft{BFGS}{\nperfprof}}\providecommand{\algeperfprof}{\StrLeft{L-BFGS-B}{\nperfprof}}\providecommand{\algfperfprof}{\StrLeft{SLSQP}{\nperfprof}}\providecommand{\alggperfprof}{\StrLeft{CMA}{\nperfprof}}\providecommand{\alghperfprof}{\StrLeft{UNIF}{\nperfprof}}\providecommand{\algiperfprof}{\StrLeft{EG50}{\nperfprof}}\providecommand{\algzeroperfprof}{best 2009}\providecommand{\perfprofsidepanel}{\mbox{\algzeroperfprof}
\vfill \mbox{\algiperfprof}
\vfill \mbox{\algfperfprof}
\vfill \mbox{\algeperfprof}
\vfill \mbox{\alggperfprof}
\vfill \mbox{\algaperfprof}
\vfill \mbox{\alghperfprof}
\vfill \mbox{\algbperfprof}
\vfill \mbox{\algcperfprof}
\vfill \mbox{\algdperfprof}}

%% file: pprldmany_05D_noiselessall.tex
\providecommand{\nperfprof}{7}\providecommand{\algaperfprof}{\StrLeft{Nelder-Mead}{\nperfprof}}\providecommand{\algbperfprof}{\StrLeft{Powell}{\nperfprof}}\providecommand{\algcperfprof}{\StrLeft{CG}{\nperfprof}}\providecommand{\algdperfprof}{\StrLeft{BFGS}{\nperfprof}}\providecommand{\algeperfprof}{\StrLeft{L-BFGS-B}{\nperfprof}}\providecommand{\algfperfprof}{\StrLeft{SLSQP}{\nperfprof}}\providecommand{\alggperfprof}{\StrLeft{CMA}{\nperfprof}}\providecommand{\alghperfprof}{\StrLeft{UNIF}{\nperfprof}}\providecommand{\algiperfprof}{\StrLeft{EG50}{\nperfprof}}\providecommand{\algzeroperfprof}{best 2009}\providecommand{\perfprofsidepanel}{\mbox{\algzeroperfprof}
\vfill \mbox{\algiperfprof}
\vfill \mbox{\alghperfprof}
\vfill \mbox{\alggperfprof}
\vfill \mbox{\algeperfprof}
\vfill \mbox{\algfperfprof}
\vfill \mbox{\algaperfprof}
\vfill \mbox{\algbperfprof}
\vfill \mbox{\algdperfprof}
\vfill \mbox{\algcperfprof}}

%% file: pprldmany_20D_separ.tex
\providecommand{\nperfprof}{7}\providecommand{\algaperfprof}{\StrLeft{Nelder-Mead}{\nperfprof}}\providecommand{\algbperfprof}{\StrLeft{Powell}{\nperfprof}}\providecommand{\algcperfprof}{\StrLeft{CG}{\nperfprof}}\providecommand{\algdperfprof}{\StrLeft{BFGS}{\nperfprof}}\providecommand{\algeperfprof}{\StrLeft{L-BFGS-B}{\nperfprof}}\providecommand{\algfperfprof}{\StrLeft{SLSQP}{\nperfprof}}\providecommand{\alggperfprof}{\StrLeft{CMA}{\nperfprof}}\providecommand{\alghperfprof}{\StrLeft{UNIF}{\nperfprof}}\providecommand{\algiperfprof}{\StrLeft{EG50}{\nperfprof}}\providecommand{\algzeroperfprof}{best 2009}\providecommand{\perfprofsidepanel}{\mbox{\algzeroperfprof}
\vfill \mbox{\algbperfprof}
\vfill \mbox{\alghperfprof}
\vfill \mbox{\algiperfprof}
\vfill \mbox{\alggperfprof}
\vfill \mbox{\algdperfprof}
\vfill \mbox{\algfperfprof}
\vfill \mbox{\algcperfprof}
\vfill \mbox{\algeperfprof}
\vfill \mbox{\algaperfprof}}

%% file: pprldmany_20D_lcond.tex
\providecommand{\nperfprof}{7}\providecommand{\algaperfprof}{\StrLeft{Nelder-Mead}{\nperfprof}}\providecommand{\algbperfprof}{\StrLeft{Powell}{\nperfprof}}\providecommand{\algcperfprof}{\StrLeft{CG}{\nperfprof}}\providecommand{\algdperfprof}{\StrLeft{BFGS}{\nperfprof}}\providecommand{\algeperfprof}{\StrLeft{L-BFGS-B}{\nperfprof}}\providecommand{\algfperfprof}{\StrLeft{SLSQP}{\nperfprof}}\providecommand{\alggperfprof}{\StrLeft{CMA}{\nperfprof}}\providecommand{\alghperfprof}{\StrLeft{UNIF}{\nperfprof}}\providecommand{\algiperfprof}{\StrLeft{EG50}{\nperfprof}}\providecommand{\algzeroperfprof}{best 2009}\providecommand{\perfprofsidepanel}{\mbox{\algzeroperfprof}
\vfill \mbox{\alggperfprof}
\vfill \mbox{\alghperfprof}
\vfill \mbox{\algiperfprof}
\vfill \mbox{\algcperfprof}
\vfill \mbox{\algdperfprof}
\vfill \mbox{\algeperfprof}
\vfill \mbox{\algfperfprof}
\vfill \mbox{\algbperfprof}
\vfill \mbox{\algaperfprof}}

%% file: pprldmany_20D_hcond.tex
\providecommand{\nperfprof}{7}\providecommand{\algaperfprof}{\StrLeft{Nelder-Mead}{\nperfprof}}\providecommand{\algbperfprof}{\StrLeft{Powell}{\nperfprof}}\providecommand{\algcperfprof}{\StrLeft{CG}{\nperfprof}}\providecommand{\algdperfprof}{\StrLeft{BFGS}{\nperfprof}}\providecommand{\algeperfprof}{\StrLeft{L-BFGS-B}{\nperfprof}}\providecommand{\algfperfprof}{\StrLeft{SLSQP}{\nperfprof}}\providecommand{\alggperfprof}{\StrLeft{CMA}{\nperfprof}}\providecommand{\alghperfprof}{\StrLeft{UNIF}{\nperfprof}}\providecommand{\algiperfprof}{\StrLeft{EG50}{\nperfprof}}\providecommand{\algzeroperfprof}{best 2009}\providecommand{\perfprofsidepanel}{\mbox{\algzeroperfprof}
\vfill \mbox{\alggperfprof}
\vfill \mbox{\algiperfprof}
\vfill \mbox{\alghperfprof}
\vfill \mbox{\algfperfprof}
\vfill \mbox{\algeperfprof}
\vfill \mbox{\algdperfprof}
\vfill \mbox{\algcperfprof}
\vfill \mbox{\algbperfprof}
\vfill \mbox{\algaperfprof}}

%% file: pprldmany_20D_multi.tex
\providecommand{\nperfprof}{7}\providecommand{\algaperfprof}{\StrLeft{Nelder-Mead}{\nperfprof}}\providecommand{\algbperfprof}{\StrLeft{Powell}{\nperfprof}}\providecommand{\algcperfprof}{\StrLeft{CG}{\nperfprof}}\providecommand{\algdperfprof}{\StrLeft{BFGS}{\nperfprof}}\providecommand{\algeperfprof}{\StrLeft{L-BFGS-B}{\nperfprof}}\providecommand{\algfperfprof}{\StrLeft{SLSQP}{\nperfprof}}\providecommand{\alggperfprof}{\StrLeft{CMA}{\nperfprof}}\providecommand{\alghperfprof}{\StrLeft{UNIF}{\nperfprof}}\providecommand{\algiperfprof}{\StrLeft{EG50}{\nperfprof}}\providecommand{\algzeroperfprof}{best 2009}\providecommand{\perfprofsidepanel}{\mbox{\algzeroperfprof}
\vfill \mbox{\alggperfprof}
\vfill \mbox{\algiperfprof}
\vfill \mbox{\alghperfprof}
\vfill \mbox{\algfperfprof}
\vfill \mbox{\algdperfprof}
\vfill \mbox{\algeperfprof}
\vfill \mbox{\algbperfprof}
\vfill \mbox{\algaperfprof}
\vfill \mbox{\algcperfprof}}

%% file: pprldmany_20D_mult2.tex
\providecommand{\nperfprof}{7}\providecommand{\algaperfprof}{\StrLeft{Nelder-Mead}{\nperfprof}}\providecommand{\algbperfprof}{\StrLeft{Powell}{\nperfprof}}\providecommand{\algcperfprof}{\StrLeft{CG}{\nperfprof}}\providecommand{\algdperfprof}{\StrLeft{BFGS}{\nperfprof}}\providecommand{\algeperfprof}{\StrLeft{L-BFGS-B}{\nperfprof}}\providecommand{\algfperfprof}{\StrLeft{SLSQP}{\nperfprof}}\providecommand{\alggperfprof}{\StrLeft{CMA}{\nperfprof}}\providecommand{\alghperfprof}{\StrLeft{UNIF}{\nperfprof}}\providecommand{\algiperfprof}{\StrLeft{EG50}{\nperfprof}}\providecommand{\algzeroperfprof}{best 2009}\providecommand{\perfprofsidepanel}{\mbox{\algzeroperfprof}
\vfill \mbox{\alggperfprof}
\vfill \mbox{\alghperfprof}
\vfill \mbox{\algeperfprof}
\vfill \mbox{\algdperfprof}
\vfill \mbox{\algcperfprof}
\vfill \mbox{\algfperfprof}
\vfill \mbox{\algiperfprof}
\vfill \mbox{\algbperfprof}
\vfill \mbox{\algaperfprof}}

%% file: pprldmany_20D_noiselessall.tex
\providecommand{\nperfprof}{7}\providecommand{\algaperfprof}{\StrLeft{Nelder-Mead}{\nperfprof}}\providecommand{\algbperfprof}{\StrLeft{Powell}{\nperfprof}}\providecommand{\algcperfprof}{\StrLeft{CG}{\nperfprof}}\providecommand{\algdperfprof}{\StrLeft{BFGS}{\nperfprof}}\providecommand{\algeperfprof}{\StrLeft{L-BFGS-B}{\nperfprof}}\providecommand{\algfperfprof}{\StrLeft{SLSQP}{\nperfprof}}\providecommand{\alggperfprof}{\StrLeft{CMA}{\nperfprof}}\providecommand{\alghperfprof}{\StrLeft{UNIF}{\nperfprof}}\providecommand{\algiperfprof}{\StrLeft{EG50}{\nperfprof}}\providecommand{\algzeroperfprof}{best 2009}\providecommand{\perfprofsidepanel}{\mbox{\algzeroperfprof}
\vfill \mbox{\alggperfprof}
\vfill \mbox{\algiperfprof}
\vfill \mbox{\alghperfprof}
\vfill \mbox{\algfperfprof}
\vfill \mbox{\algdperfprof}
\vfill \mbox{\algeperfprof}
\vfill \mbox{\algcperfprof}
\vfill \mbox{\algbperfprof}
\vfill \mbox{\algaperfprof}}